\documentclass[10pt,journal,compsoc]{IEEEtran}

\usepackage{eurosym}
\usepackage{amsfonts}
\usepackage[nocompress]{cite}
\usepackage{cite}
\usepackage{amssymb}
\usepackage{amsmath}
\usepackage{amsthm}
\usepackage{graphicx}
\usepackage{times}
\usepackage{epsfig}
\usepackage{subcaption}
\usepackage{booktabs}
\usepackage[pagebackref=true,breaklinks=true,letterpaper=true,colorlinks,bookmarks=false]{hyperref}

\ifCLASSOPTIONcompsoc
\else
\fi
\ifCLASSINFOpdf
\else
\fi
\input{tcilatex}
\begin{document}

\title{FSGANv2: Improved Subject Agnostic\\Face Swapping and Reenactment}
\author{Yuval~Nirkin, Yosi~Keller, and~Tal~Hassner
\IEEEcompsocitemizethanks{\IEEEcompsocthanksitem Y. Nirkin and Y. Keller are with the Faculty of Engineering. Bar Ilan University.
\and
E-mail: yuval.nirkin@gmail.com
\IEEEcompsocthanksitem Tal Hassner E-mail: talhassner@gmail.com}\thanks{%
This paper reports work done entirely at Bar-Ilan University, Israel.}}
\maketitle

\begin{abstract}
We present Face Swapping GAN (FSGAN) for face swapping and reenactment.
Unlike previous work, we offer a subject agnostic swapping scheme that can
be applied to pairs of faces without requiring training on those faces.
We derive a novel iterative deep learning--based approach for face
reenactment which adjusts significant pose and expression variations that
can be applied to a single image or a video sequence. For video sequences,
we introduce a continuous interpolation of the face views based on
reenactment, Delaunay Triangulation, and barycentric coordinates. Occluded
face regions are handled by a face completion network. Finally, we use a
face blending network for seamless blending of the two faces while
preserving the target skin color and lighting conditions. This network uses
a novel Poisson blending loss combining Poisson optimization with a
perceptual loss. We compare our approach to existing state-of-the-art
systems and show our results to be both qualitatively and quantitatively
superior. This work describes extensions of the FSGAN method, proposed in an earlier conference version of our work~\cite{nirkin2019fsgan}, as well as additional experiments and results.
\end{abstract}

\markboth{}{Shell
\MakeLowercase{\textit{et al.}}: Bare Demo of IEEEtran.cls for Computer Society Journals}


\section{Introduction}

\label{sec:introduction}

\begin{figure*}[!htb]
\centering
\includegraphics[width=1.0\textwidth]{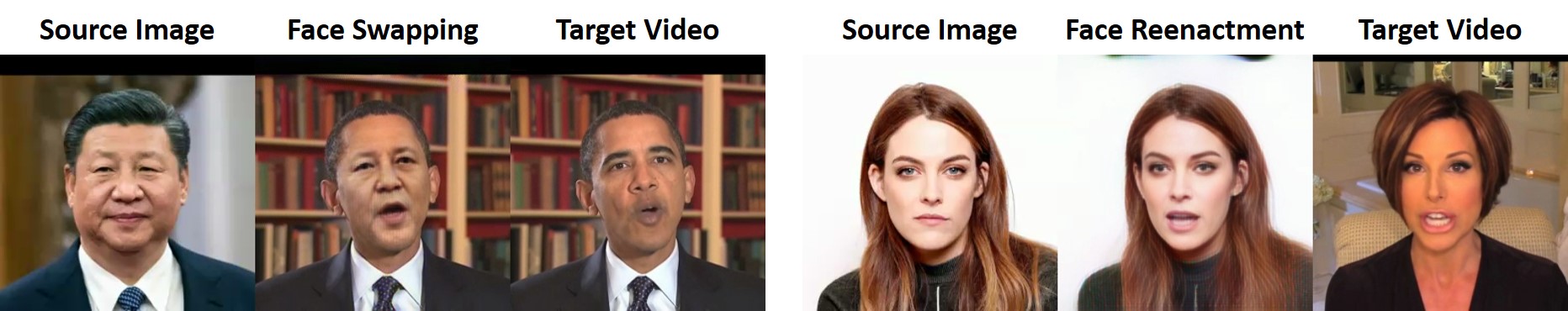}
\caption{\emph{Face swapping and reenactment.} Left: Source face swapped
onto target. Right: Target video used to control the expressions of the face
appearing in the source image. In both cases, our results appear in the
middle. }
\label{fig:teaser}
\end{figure*}

\IEEEPARstart{F}{ace swapping} is the visual transformation of a face from a
source image to a target image, such that, the resulting image seamlessly
replaces the face appearing in the target image as depicted in Fig.~\ref%
{fig:teaser}. \emph{Face reenactment} (aka \emph{face transfer} or \emph{%
puppeteering}) utilizes the facial movements and expression of a control
face in one video to guide the motions and deformations of a second face
appearing in another video or image (Fig.~\ref{fig:teaser}). Both tasks have
attracted significant research attention due to their applications in
entertainment~\cite%
{alexander2009creating,kemelmacher2016transfiguring,wolf2010eye}, privacy~%
\cite{blanz2004exchanging,lin2012face,mosaddegh2014photorealistic}, 
and training data generation and augmentation, specifically for detection of such methods~\cite{nirkin2020deepfake,masi2020two,guera2018deepfake}.
Most contemporary works proposed methods for either swapping or reenactment, but rarely both, relying on
underlying 3D face representations to transfer the face appearance. Face
shapes were estimated from the input image~\cite%
{Hassner:ICCV13:ViewFace3D,thies2016face2face,suwajanakorn2017synthesizing,masi2017rapid,nirkin2018face} or kept
fixed~\cite{hassner2015effective,nirkin2018face,masi2019face}. The 3D shape was then aligned with the input
images and used as a proxy to transfer the image appearance (swapping) or
controlling the facial expression and viewpoint (reenactment).

Deep networks were also applied to face swapping and reenactment. In
particular, Generative Adversarial Networks (GANs)~\cite%
{goodfellow2014generative} were shown to successfully generate realistic
fake faces images. Conditional GANs (cGANs)~\cite%
{mirza2014conditional,isola2017image,wang2018pix2pixHD} were used to
transform an image depicting real data from one domain to the other and
inspired multiple face reenactment schemes~\cite%
{pumarola2018ganimation,wayne2018reenactgan,sanchez2018triple}. The
DeepFakes project~\cite{DeepFakes} applied cGANs to face swapping in videos,
making swapping widely accessible to nonexperts and receiving substantial
public attention. Such methods are able to generate more realistic face
images by replacing the classical graphics pipeline by utilizing implicit 3D
face representations.

Some methods applied domain separation~in latent feature spaces \cite%
{tian2018cr,natsume2018rsgan,natsume18fsnet}, to decompose the identity
component of a face from the other traits, such as pose and expression. The
identity is encoded as the manifestation of latent feature vectors,
resulting in significant information loss and limiting the quality of the
synthesized images. Subject-specific approaches \cite%
{suwajanakorn2017synthesizing,DeepFakes,wayne2018reenactgan,kim2018deep} are
particularly trained for each subject or pair of subjects to be swapped or
reenacted. Thus, requiring significant training sets per subject, to achieve
reasonable results, limiting their potential usage. A major concern shared
by previous face synthesis schemes, particularly the 3D-based methods, is
that they all require particular care to handle partially occluded faces.

In this work, we improve upon our previous work~\cite{nirkin2019fsgan} in multiple ways. We provide a means for interpolating
between face landmarks without relying on 3D information using a face landmarks transformer network. We improve the inpainting generator by adding symmetry and face landmarks cues. We completely revise the preprocessing pipeline and add a postprocessing step, to reduce the jittering and saturation artifacts of our previous method. Finally, we show additional qualitative and quantitative experiments with a new metric for comparing expressions.

To summarize, our contributions are:
\begin{itemize}
\item A face landmarks transformer network for interpolating between face landmarks without 3D information.
\item Improved inpainting generator that utilizes symmetry and face landmarks cues.
\item A demonstration of an additional use case for the new face reenactment method for pose-only face reenactment.
\item Completely revised preprocessing and an additional postprocessing step for reducing hittering and saturation artifacts.
\item Introduction of a new metric for facial expression comparison.
\item Additional quantitative and qualitative experiments and ablation studies using new metrics.
\end{itemize}

\section{Related work}

\label{sec:related_work}

Methods for manipulating the appearance of face images, particularly for
face swapping and reenactment, were originally proposed to address privacy
concerns~\cite{blanz2004exchanging,lin2012face,mosaddegh2014photorealistic},
though they are increasingly being used for recreational~\cite%
{kemelmacher2016transfiguring} or entertainment tasks (e.g.,~\cite%
{alexander2009creating,wolf2010eye}).

\subsection{3D-based methods}

One of the earliest face swapping methods \cite{blanz2004exchanging}
required manual involvement, where a fully automatic method was
later~proposed \cite{bitouk2008face}. The Face2Face approach transferred
expressions from the source to target faces~\cite{thies2016face2face}, by
fitting a 3D morphable face model (3DMM)~\cite{blanz2002face,blanz2003face}
to both faces. The expression attributes of one face were then applied to
the other, where particular attention was given to the interior mouth
regions. The reenactment method of Suwajanakorn et al.~\cite%
{suwajanakorn2017synthesizing} synthesized the mouth area using a
reconstructed 3D model of (former president) Barack Obama, guided by face
landmarks, and using a similar approach for filling the face interior as in
Face2Face. The expression of frontal faces was manipulated by Averbuch-Elor
et al.~\cite{averbuch2017bringing} by transferring the mouth interior from
source to the target image using 2D wraps and face landmarks.

Finally, Nirkin et al.~\cite{nirkin2018face} proposed a face swapping
method, demonstrating that 3D face shape estimation is unnecessary for
realistic face swaps. Instead, they used a fixed 3D face shape as the proxy.
This approach, similar to ours, proposed a face segmentation method, though
the scheme was not end-to-end trainable and required particular attention to
occluded regions. We show our current results to be superior.

\subsection{GAN-based methods}

\begin{figure*}[!htb]
\centering
\includegraphics[width=1.0\textwidth]{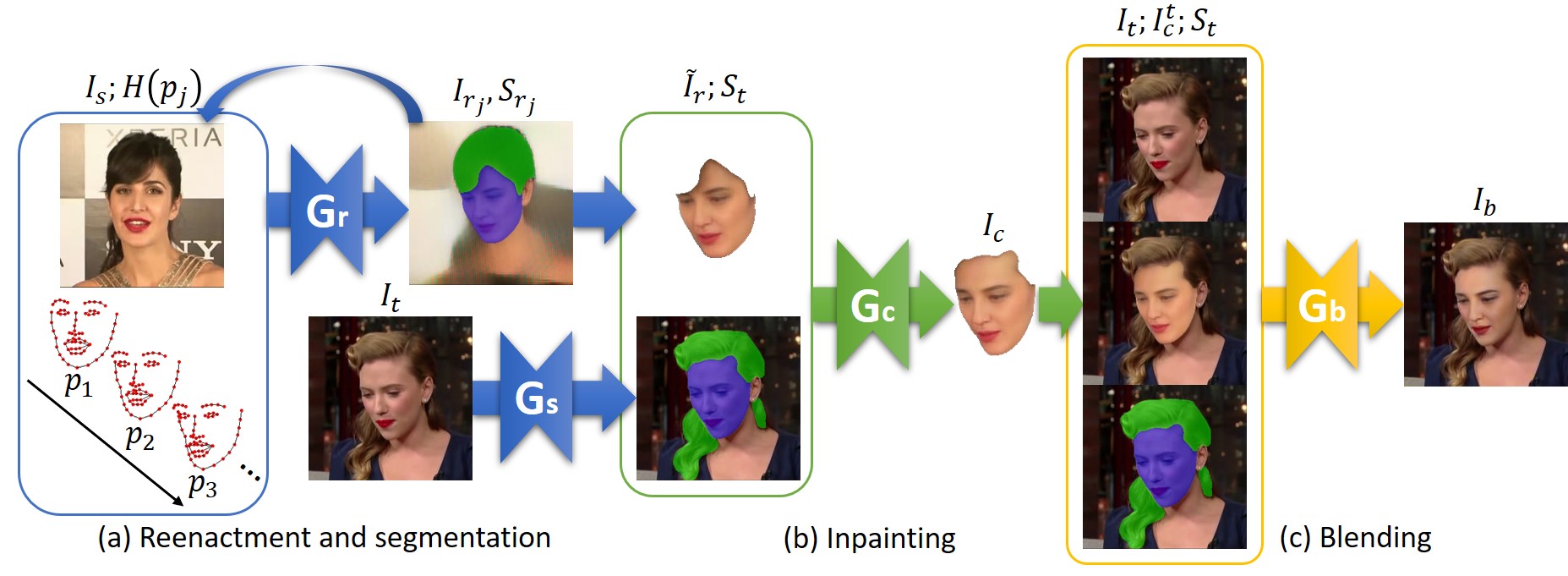}
\caption{Overview of the proposed FSGAN approach. (a) The recurrent
reenactment generator, $G_{r}$, and the segmentation generator, $G_{s}$. $G_{r}$
estimates the reenacted face, $F_{r}$, while $G_{s}$ estimates the face and
hair segmentation mask, $S_{t}$, of the target image, $I_{t}$. (b) The
inpainting generator, $G_{c}$, inpaints the missing parts of $\tilde{F}_{r}$
based on $S_{t}$ to estimate the completed reenacted face, $F_{c}$. (c) The
blending generator, $G_{b}$, blends $F_{c}$ and $F_{t}$, using the
segmentation mask, $S_{t}$. }
\label{fig:system}
\end{figure*}

GANs~\cite{goodfellow2014generative} were shown to generate fake images
having the same distribution as the target domain. Although successful in
generating realistic appearances, their training might be unstable, thus
limiting their applicability to generating low-resolution images. Subsequent
methods, however, improved the stability of the training process~\cite%
{mao2017least,arjovsky2017wasserstein}. Karras et al.~\cite%
{karras2017progressive} train GANs using a progressive coarse-to-fine
multiscale scheme. CycleGAN~\cite{zhu2017unpaired} proposed a cycle
consistency loss, allowing to train unsupervised generic transformations
between different domains. A cGAN with $L_{1}$ loss was applied by Isola et
al.~\cite{isola2017image} to derive the pix2pix method, and was shown to
produce realistic image synthesis results for applications such as
transforming edges to faces.

Recently, attention-based upsampling was proposed to improve
generation and dense prediction, to reduce the generation artifacts and
improve the prediction accuracy~\cite{wang2019carafe,lu2019indices}.

\subsection{Facial manipulation using GANs}

GANs were used for high resolution image-to-image translation by pix2pixHD~%
\cite{wang2018pix2pixHD}, that applied a multiscale cGAN architecture, and
a perceptual loss~\cite{johnson2016perceptual} using a pre-trained VGG
network that improved the results. GANimation~\cite{pumarola2018ganimation}
proposed a dual generator cGAN architecture conditioned on emotion action
units that generate an additional attention map. This map was used to
interpolate between the reenacted and original images to preserve the
background. The GANnotation~\cite{sanchez2018triple} deep facial reenactment
scheme was driven by face landmarks, where the images were generated
progressively using a triple consistency loss. The images were first
frontalized using landmarks before further processing.

A hybrid 3D/deep method was proposed by Kim et al.~\cite{kim2018deep}, that
render a reconstructed 3DMM of a specific subject using a classical graphic
pipeline. The rendered image is then processed by a generator network,
trained to map synthetic rendered views of each subject to photo-realistic
images. Finally, feature disentanglement was proposed for face manipulation.
RSGAN~\cite{natsume2018rsgan} disentangles the latent face and hair
representations, while FSNet~\cite{natsume18fsnet} proposed a latent space
which separates the identity and geometric attributes, such as facial pose
and expression.

Zakharov et al.~\cite{zakharov2019few} derived a few-shot face
reenactment approach comprised of two parts: a meta-learning stage where the
networks are trained using the face images of multiple subjects, and a
task-specific phase, where the network is fine-tuned using the images of a
particular person. In both stages, an encoder network learns to extract
identity cues, and a generator CNN learns to map this information alongside
the face landmarks to generate novel views.

Video dubbing~\cite{kim2019neural,thies2019neural} was applied using visual and audio
information, respectively. The first used a recurrent GAN to preserve the
expression style of the target. The second, extracted deep audio features
which were encoded as expression coefficients. To generate the final
rendering, both methods used neural rendering to translate classic
renderings into realistic ones using an additional person-specific
generator.
Another kind of video dubbing was performed by text editing~\cite{thies2019neural,fried2019text,yao2020iterative}, Fried et al.~\cite{fried2019text} by mapping input text to 3DMM pose and expression which was used to recurrently render the lower part of the face, and Yao et al.~\cite{yao2020iterative} suggested an interactive text editing using fast phoneme matching and neural rendering.

Ha et al.~\cite{ha2019marionette} suggested a face reenactment
framework using a few-shot setting with a landmark transformer that adjusts
the shape and blending components using an image attention block. Li et al.~%
\cite{li2019faceshifter} introduced a face swapping method that preserves
the attributes of the source face by integrating them with the generator in
a multi-scale scheme using attention blocks. In addition, occlusions are
handled in an unsupervised way by comparing the reconstructed and target
images.

Recently, Naruniec et al.~\cite{naruniec2020high} proposed a face swapping method at megapixel resolution using an encoder-decoder with contrast and light-preserving blending and landmark stabilization. Zakharov et al.~\cite{zakharov2020fast} introduced a one-shot neural rendering-based face reenactment method that can perform in real-time. They achieve this by first rendering a coarse image and then combining it with a warped texture image that is generated offline.

\section{Face swapping GAN}

\label{sec:FSGAN}

\begin{figure*}[ptb]
\centering
\includegraphics[width=\linewidth]{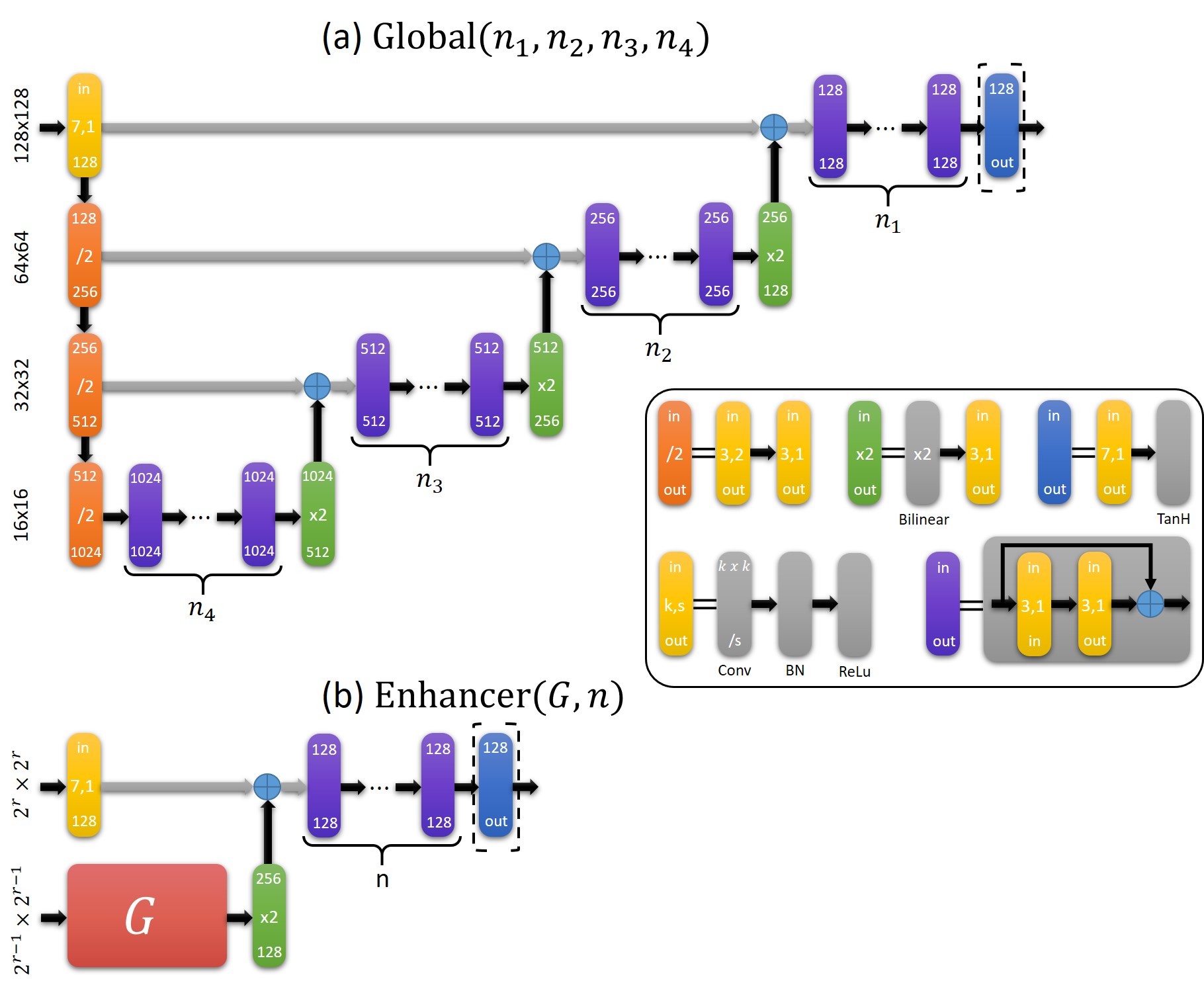}
\caption{ \emph{Generator architectures.} (a) The global generator is based
on a residual variant of the U-Net~\protect\cite{ronneberger2015u} CNN,
using a number of bottleneck layers per resolution. We replace the simple
convolutions with bottleneck blocks (in purple), the concatenation with
summation (plus sign), and the deconvolutions with bilinear upsamplnig
following by a convolution. (b) The enhancer utilizes a submodule and a
number of bottleneck layers. The last output block (in blue) is only used in
the enhancer of the finest resolution.}
\label{fig:architecture}
\end{figure*}

We introduce the Face Swapping GAN (FSGAN) whose overview is
illustrated in Fig.~\ref{fig:system}. Let $I_{s}$ and $I_{t}$ be the source and
target face images, respectively, representing the faces $F_{s}\in I_{s}$ and $%
F_{t}\in I_{t}$. We aim to synthesize a new image based on $I_{t}$, such
that $F_{t}$ is seamlessly replaced by $F_{s}$ while retaining the same pose
and expression.

The FSGAN consists of three major components. The first, detailed in Section~%
\ref{subsec:reenactment} and shown in Fig.~\ref{fig:system}(a), consists of
the reenactment generator, $G_{r}$, and the segmentation CNN, $G_{s}$. The
input to $G_{r}$ are the heatmaps encoding the facial landmarks of $F_{t}$,
and generates the reenacted image, ${I}_{r}$, such that $F_{r}$ denotes $%
F_{s} $ having the same pose and expression as $F_{t}$. It also computes $%
S_{r}$; that is, the segmentation mask of $F_{r}$. The $G_{s}$ component
computes the segmentations mask of the face and hair of $F_{t}$.

Given the reenacted image, $I_{r}$, there might be missing face parts, as
demonstrated in Figs.~\ref{fig:system}(a) and \ref{fig:system}(b). Hence, we
apply the face inpainting network, $G_{c}$, detailed in Section~\ref%
{subsec:inpainting} using the segmentation, $S_{t}$, to inpaint the missing
parts. The final phase of the FSGAN, shown in Fig.~\ref{fig:system}(c) and
detailed in Section~\ref{subsec:Blending}, is the blending of the completed
face, $F_{c}$, into the target image, $I_{t}$, to derive the final face swapping
result.

The architecture of our face segmentation network, $G_{s}$, is based on the
U-Net~\cite{ronneberger2015u}, with bilinear interpolation for upsampling,
while all other generators---$G_{r}$, $G_{c}$, and $G_{b}$---are based on
the pix2pixHD architecture~\cite{wang2018pix2pixHD}, utilizing a
coarse-to-fine generator, and a multi-scale discriminator.

For the global generator, we use a U-Net with bottleneck blocks~\cite%
{he2016deep} instead of simple convolutions, and summation instead of
concatenation. The summation allows for smaller channel dimensions than by
using concatenation, while bottleneck blocks improve the convergence of
large generator models \cite{MarioNETte:AAAI2020}. As with the segmentation
network, we use bilinear interpolation for upsampling in both the global
generator and the enhancers. The actual number of layers differs per each
generator, and the architectures are detailed in Section~\ref%
{subsec:architecture}.

Following previous results \cite{wayne2018reenactgan}, training
subject-agnostic face reenactment is non-trivial and might fail when applied
to face images related by large poses. For that, we propose to break-down
large pose changes into smaller reenactment steps and interpolate between
the closest available source images corresponding to the target's pose.
These steps are detailed in the following sections.

\subsection{Detection and tracking}

\label{subsec:Detection-tracking}

Each video is first processed by the dual-shot face detector (DSFD)~%
\cite{li2019dsfd} that is more accurate than SFD~\cite{zhang2017s3fd}, which was used in our previous work. The detections are grouped into sequences by computing
the IoUs of the detections in successive frames, where IoUs \TEXTsymbol{>} 
0.75 are grouped together in the same sequence. 
The facial expressions were tracked using 2D face landmarks \cite{sun2019deep},
trained on the WLFW dataset, consisting of 98 points per face.
In our previous work, we used 2D and 3D landmarks of 68 points~\cite{bulat2017far}.

Even when using state-of-the-art face detection and landmarks
extraction schemes, the bounding boxes and landmarks point are not sub-pixel
accurate, causing inconsistencies between subsequent frames, resulting in
noticeable jittering, an apparent issue of our previous method. Temporal averaging might help reduce the jittering but
introduces a noticeable lag. For that, we extend the 1\euro\ filter~\cite%
{casiez20121}, based on the fact that humans are more sensitive to jittering added to small
motions, and more sensitive to lags, when there is a large motion.

We estimate the motion following Casiez et al. \cite{casiez20121},
but use it as per-frame weights for a 1D temporal averaging filter. The
larger the motion, the less averaging is applied. For the bounding boxes, we
apply the smoothing to the center and box dimensions separately, and for
the face landmarks we apply the smoothing to each face part separately.

\subsection{Generator architecture}

\label{subsec:architecture}

The generators $G_{r}$, $G_{c}$, and $G_{b}$ follow the pix2pixHD design, as depicted in Fig.~\ref{fig:architecture}. 
The global generator, $\text{Global}(n_{1},n_{2},n_{3})$, 
is defined with respect to the number of bottleneck blocks $n_{1},n_{2},n_{3}$ (shown in purple), corresponding to each of the three
scales. The enhancer, $\text{Enhancer}(G,n)$, is defined with respect to its
submodule and $n$ bottleneck blocks. Our reenactment generator and
completion generators are%
\begin{equation}
G_{r}=G_{c}=Enhancer(Global(2,2,3),2),
\end{equation}%
and the blending generator is%
\begin{equation}
G_{b}=Enhancer(Global(1,1,1),1).
\end{equation}%

\subsection{Training losses}

\label{subsec:Training-Losses}

\begin{figure*}[tbh]
\centering
\includegraphics[width=0.9\textwidth]{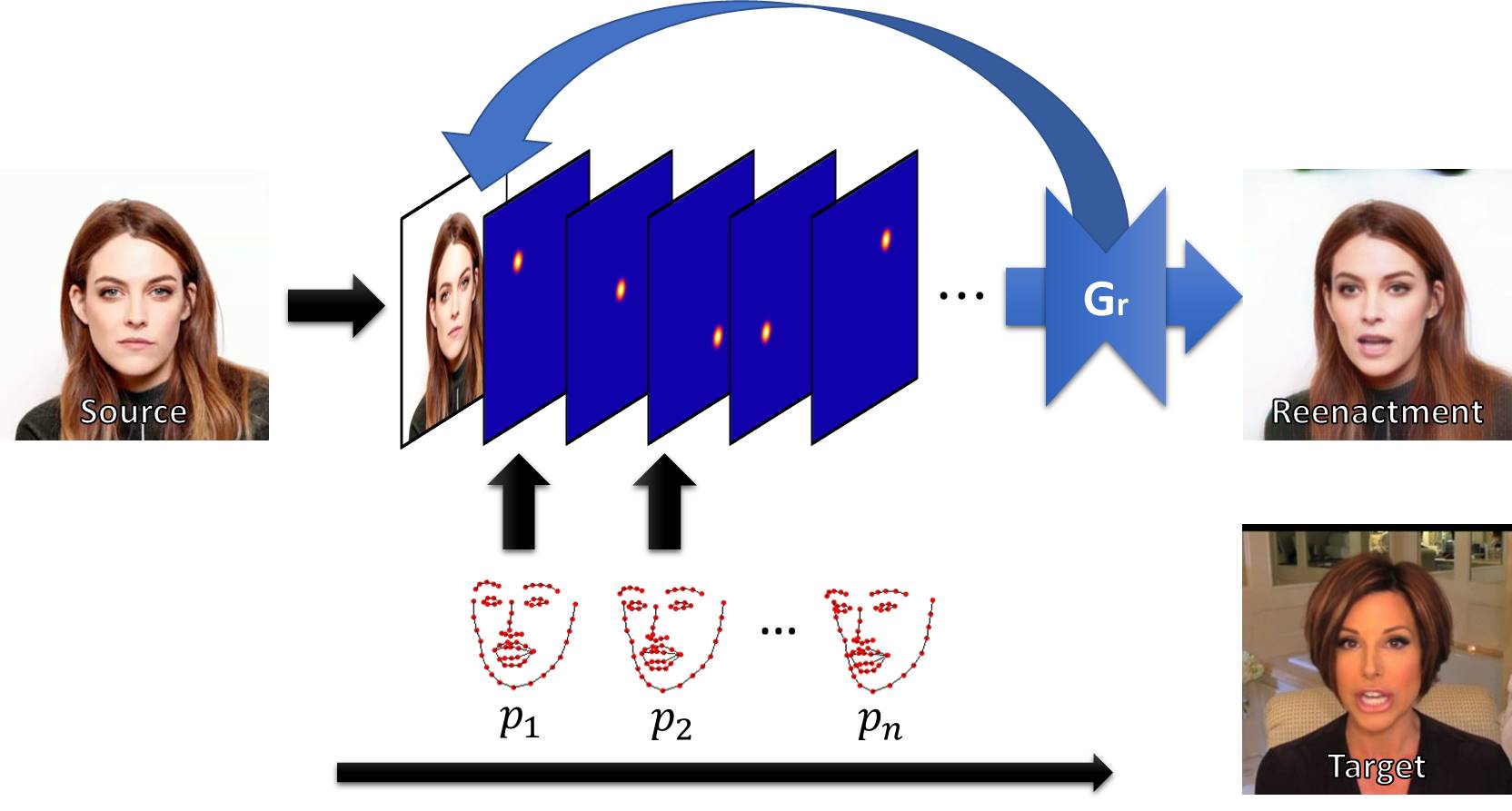}
\caption{\emph{Face reenactment.} A single iteration of our proposed face
reenactment method. The source image is concatenated with the heatmap
corresponding to the first face landmarks, $p_{1}$, and fed to the
reenactment generator, $G_{r}$, producing the first reenactment
iteration result. The next iteration uses the result of the previous
iteration instead of the source image with the next face landmarks.}
\label{fig:reenactment}
\end{figure*}

\subsubsection{Domain specific perceptual loss}

\label{subsec:Perceptual Loss}

The perceptual loss \cite{johnson2016perceptual}, that has been widely used
in recent works in face synthesis~\cite{sanchez2018triple}, outdoor scenes~%
\cite{wang2018pix2pixHD}, and super resolution~\cite{ledig2017photo}, is
applied to capture the fine details of the face texture.~A feature maps of a
pretrained VGG network is used to compare the high frequency details using a
Euclidean distance.
Following others~\cite{zhang2018unreasonable,tej2020enhancing}, who show that the effectiveness of perceptual loss can be improved by fine-tuning on specific domains, we propose to train the VGG-19 CNN \cite{simonyan2014very} used for perceptual loss using multiple face
recognition and attribute classification datasets, instead of a generic
ImageNet-based\ CNN. This allows to fully capture the distinct details
inherent in face images. Let $F_{i}\in
\mathbb{R}
^{C_{i}\times H_{i}\times W_{i}}$ be the feature map of the $i$-th layer of
the VGG-19 network, the perceptual loss is thus given by%
\begin{equation}
\mathcal{L}_{perc}(x,y)=\sum_{i=1}^{n}\frac{1}{C_{i}H_{i}W_{i}}\left\Vert
F_{i}(x)-F_{i}(y)\right\Vert _{1}
\end{equation}
where $C_i$ is the number of channels, and $H_i$ and $W_i$ are the height and width dimensions.

\subsubsection{Reconstruction loss}

\label{subsubsec:Reconstruction-Loss}

While the perceptual loss, as in Section \ref{subsec:Perceptual Loss},
captures the high frequencies well, generators trained using only that loss,
often produce images having inaccurate colors, corresponding to the
erroneous reconstruction of the low-frequency image content. Hence, we also
applied a pixelwise $L_{1}$ loss to the generator%
\begin{equation}
\mathcal{L}_{pixel}(x,y)=\Vert x-y\Vert _{1},
\end{equation}%
and the overall loss is given by
\begin{equation}
\mathcal{L}_{rec}(x,y)=\lambda _{perc}\mathcal{L}_{perc}(x,y)+\lambda
_{pixel}\mathcal{L}_{pixel}(x,y)  \label{equ:total loss}
\end{equation}

The loss in Eq. \ref{equ:total loss} was applied to all image generators.

\subsubsection{Adversarial loss}

\label{subsubsec:Adversarial-Loss}

To further improve the realism of the synthesized images, we applied the
pix2pixHD adversarial objective \cite{wang2018pix2pixHD}, which utilizes a
multi-scale discriminator comprising of multiple discriminators, $D_{1},D_{2},...,D_{n}$, 
each operating on a different image resolution. For
a generator, $G$, and a multi-scale discriminator, $D$, our adversarial loss is
\begin{equation}
\mathcal{L}_{adv}(G,D)=\min_{G}\max_{D_{1},\dots D_{n}}\sum_{i=1}^{n}%
\mathcal{L}_{GAN}(G,D_{i}),
\end{equation}%
where
\begin{align}
\mathcal{L}_{GAN}(G,D)=& \mathbb{E}_{(x,y)}[\log D(x,y)]  \notag \\
& +\mathbb{E}_{x}[\log (1-D(x,G(x)))].
\end{align}

\subsection{Face Segmentation}

\label{subsec:segmentation}

The goal of our face segmentation is to distinguish between the
internal part of the face and the background, including the rest of the
head. In practice, we found that adding information regarding the hair
improves the segmentation accuracy. Given an image , $I\in R^{3\times H\times
W}$, we define the face segmentation generator, $G_{s}:R^{3\times
H\times W}\rightarrow R^{3\times H\times W}$. The output
segmentation mask predicts three classes: background, face, and hair.
$G_s$ was trained using the cross-entropy loss.
To improve stability and accuracy, the training images are augmented by
random rotations of 
$\left[ \mathbf{-30{{}^\circ},30{{}^\circ}}\right]$, random color jittering (brightness, contrast,
saturation, and hue), random horizontal flip, and random gaussian blur with
a kernel radius of 5 and standard deviation of 1.1.

\subsection{Face reenactment}

\label{subsec:reenactment}

As in~\cite{nirkin2019fsgan}, the purpose of the face reenactment generator, $G_r$, is to align a \textit{source} face according to the pose and expression of a \textit{target} face, without subject-specific training. Similarly to~\cite{nirkin2019fsgan}, significant poses are decomposed into smaller, more manageable steps, in which $G_r$ is applied iteratively, but instead of relying on 3D landmarks, we propose a landmarks transformer network to perform the interpolation between the landmarks in each step, with more details on this in Section~\ref{subsubsec:landmarks_transformer}. A single reenactment
iteration is depicted in Fig~\ref{fig:reenactment}.

Additionally, the newly proposed $G_r$ benefit from improved preprocessing, which allows for extracting higher quality cues from the same training videos, a more accurate landmarks extraction method~\cite{sun2019deep}, and improved heatmap encoding method that is discussed in Section~\ref{subsubsec:landmarks_heatmaps}.

We use an appearance map of the source face to identify and utilize
the nearest available views, as detailed in Section~\ref%
{subsec:FaceViewInterpolation}. Given an image, $I\in R^{3\times H\times W}$,
 and a heatmap representation, $H(p)\in R^{98\times H\times W}$,
 of the facial landmarks, $p\in R^{98\times 2}$, we define the face reenactment generator
\begin{equation}
\mathbf{\ }G_{r}:\left\{ \mathbb{R}^{3\times H\times W},\mathbb{R}^{98\times
H\times W}\right\} \rightarrow R^{3\times H\times W}.
\end{equation}%
Let $p_{s},p_{t}\in R^{98\times 2}$ and $\theta
_{s},\theta _{t}\in R^{3}$, be the 2D landmarks and Euler angles of %
$F_{s}$ and $F_{t}$, respectively. Utilizing $n$
reenactment iterations, we generate $n-1$ intermediate 2D
landmarks, $p_{1},\dots ,p_{n-1}$:
\begin{align}
& p_{i}=T\left( p_{s},\left( 1-\frac{i}{n}\right) \theta _{s}+\frac{i}{n}\theta _{t}\right) \qquad \text{for }i=1,\dots ,n-1, \\
& p_{n}=p_{t}.  \notag
\end{align}

where $T$ is the landmark transformer discussed in the next section. The recursive reenactment output of each iteration is thus given by
\begin{align}
& I_{r_{i}}=G_{r}(I_{r_{i-1}};H(p_{i})),1\leq i\leq n \\
& I_{r_{0}}=I_{s}.  \notag
\end{align}

\subsubsection{Landmarks transformer}
\label{subsubsec:landmarks_transformer}

To generate the intermediate face landmarks we need to rotate the source landmarks by the delta angle. In our previous work \cite{nirkin2019fsgan}, 3D landmarks were used with geometric transformations, in this work we rely on 2D landmarks and pose estimation, and perform the transformation using a CNN. The 2D landmarks are
detected as in Sun et al. \cite{sun2019deep}, while the pose is estimated
using landmark free pose estimation method~\cite{ruiz2018fine,chang2017faceposenet}.

Given the landmarks, $p_{s}\in R^{98\times 2}$, in the source image, and the head pose, $\theta_{t}\in R^{3}$, in the target image, we aim to estimate the 2D
landmarks, $p_{t}$, in the target image. Thus, we propose the
landmarks transformer, $T:\left\{ p_{1},\theta _{1}\right\} \rightarrow p_{2}
$. The network is comprised of 12 linear layers, each followed by
batch normalization and ReLU activation layers except for the final layer.
It is trained using sequences of faces, where we randomly select a pair of
face images, $F_{s}$ and $F_{t}$, and minimize the mean
square error between the predicted landmarks, $T(p_{s_{i}},\theta_{t_{i}})$, and the ground truth target landmarks, $p_{t_{i}}$:%
\begin{equation}
\Vert x-y\Vert _{1}
\mathcal{L}(T)=\frac{1}{n}\sum_{i=1}^{n}\Vert T(p_{s_{i}},\theta_{t_{i}})-p_{t_{i}}\Vert_{2}^{2}.
\end{equation}

\subsubsection{Landmarks heatmaps}
\label{subsubsec:landmarks_heatmaps}

\begin{figure*}[t]
\centering
\includegraphics[clip,trim=0mm 3mm 0mm 0mm,
width=.85\textwidth]{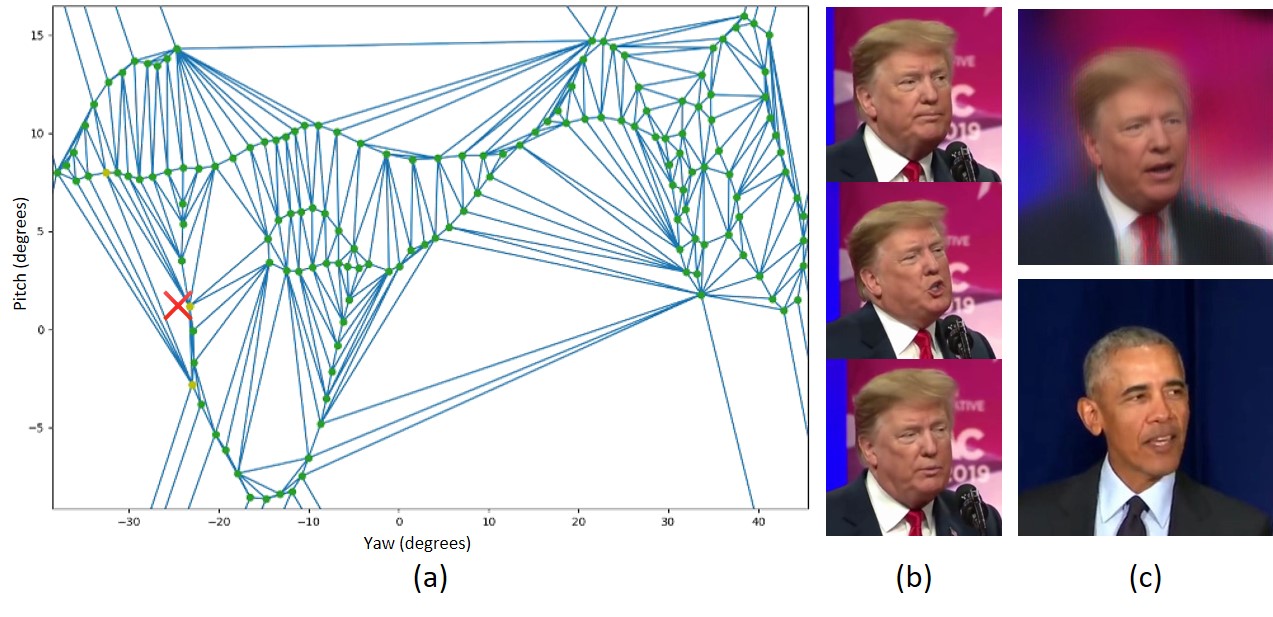}\vspace{-3mm}
\caption{ \emph{Face view interpolation.} (a) Shows an example of an
appearance map of the source subject (Donald Trump). The green dots
represent different views of the source subject, the blue lines represent
the Delaunay Triangulation of those views, and the red X marks the location
of the current target's pose. (b) The interpolated views associated with the
vertices of the selected triangle (represented by the yellow dots). (c) The
reenactment result and the current target image. \protect\vspace{-4mm} }
\label{fig:interpolation}
\end{figure*}

Let $\mathbf{A}\in R^{C\times H\times W}$ be the output
layer of the face landmarks network~\cite{sun2019deep}, where $C$
is number of landmark points. 
In our previous work, the face landmarks were encoded by taking the points of maximum activation in each channel in $A$, and were decoded by splatting the points on the heatmap followed by a Guassian blur filter. Both operation are not sub-pixel accurate. To further reduce jittering and improve temporal consistency, we replace those two operations by sub-pixel accurate encoding and decoding.

We filter-out the lower values in $%
\mathbf{A}$ corresponding to non-landmarks by normalizing $\mathbf{%
A}$ to $[0,1]$ and zeroing $\mathbf{A<0.5}$. A
landmark point, $p_{c}$, encoded by the channel, $C$, in $%
\mathbf{A}$, is given by%
\begin{equation}
p_{c}=\sum_{x,y}\mathbf{A}(C,y,x)\cdot (x+0.5,y+0.5),
\end{equation}%
After encoding the face landmarks we perform temporal
smoothing as in Section~\ref{subsec:Detection-tracking}. The heatmap layer
corresponding to $p$ is given by
\begin{equation}
\widetilde{H}(p)_{c,y,x}=1-\frac{1}{\sqrt{2}}\left\Vert
p_{c}-(x+0.5,y+0.5)\right\Vert _{2},
\end{equation}%
and the support of a heatmap point is given by
\begin{equation}
H(p)=\max \left( \frac{\widetilde{H}(p)^{8}-t}{1-t},0\right) ,
\end{equation}%
where $t=0.8$ was used in all our experiments.

\subsubsection{Training}

\label{subsubsec:reenactment_training}

Denote by $I_{s},I_{t}$, the source and target images
denoting the same subject, where both images are randomly selected from the
same video sequence. Indicated by $I_{r_{n}}$, the reenactment result
after $n$ iterations, and $I_{r}$, the reenactment results without
intermediate iterations. $\widetilde{I}$ is image $I$, 
but with its background removed. Thus, the loss of the generator, $G_{r}$, is
\begin{align}
\mathcal{L}(G_{r})=& \lambda _{stepwise}\mathcal{L}_{rec}(\widetilde{I}%
_{r_{n}},\widetilde{I}_{t})+\lambda _{rec}\mathcal{L}_{rec}(\widetilde{I}%
_{r},\widetilde{I}_{t})  \notag \\
& +\lambda _{adv}\mathcal{L}_{adv},
\end{align}%
where $L_{rec}$ is given in Eq. \ref{equ:total loss}.

\subsection{Face view interpolation}

\label{subsec:FaceViewInterpolation}

Following the common computer graphics pipeline~\cite{hughes2014computer},
in which textured mesh polygons are projected onto a plane and seamlessly
rendered, we propose a novel scheme for iterative interpolation between face
views. This step is an essential phase of our method, as it enables us to
utilize the entire source video sequence, without training our model on a
particular video frame, thus allowing our approach to be subject agnostic.

Given a set of source subject images, $\left\{ \mathbf{I}_{s_{1}},\dots ,%
\mathbf{I}_{s_{n}}\right\}$, and Euler angles, $\left\{ \mathbf{e}_{1},\dots
,\mathbf{e}_{n}\right\}$, of the corresponding faces, $\left\{ \mathbf{F}%
_{s_{1}},\dots ,\mathbf{F}_{s_{n}}\right\}$, we construct the appearance map
of the source subject illustrated in Fig.~\ref{fig:interpolation}(a). The
appearance map embeds the head poses in a triangulated plane, allowing to
follow continuous paths of head poses.

We start by projecting the Euler angles, $\left\{ \mathbf{e}_{1},\dots ,%
\mathbf{e}_{n}\right\}$, to a plane by dropping the roll angle. Using a k-d
tree data structure~\cite{hughes2014computer}, we prune points in the
angular domain that are too close to each other. We retain points for which
the corresponding Euler angles have a roll angle closer to zero, and remove
motion blurred images. Using the remaining points, $\left\{ x_{1},\dots
,x_{m}\right\}$, and the four boundary points, $y_{i}\in \lbrack
-75,75]\times \lbrack -75,75]$, we build a mesh $M$ in the angular domain
using a Delaunay Triangulation.

For a query Euler angle $e_{t}$, of a face, $F_{t}$, and its corresponding
projected point, $x_{t}$, we find the triangle, $T\in M$, that contains $x_{t}$%
. Let $x_{i_{1}},x_{i_{2}},x_{i_{3}}$ be the vertices of $T$ and $%
I_{s_{i_{1}}},I_{s_{i_{2}}},I_{s_{i_{3}}}$ be the corresponding face views.
We calculate the barycentric coordinates, $\lambda _{1},\lambda _{2},\lambda
_{3}$, of $x_{t}$, with respect to $x_{i_{1}},x_{i_{2}},x_{i_{3}}$. The
interpolation result, $I_{r}$, is then given by

\begin{equation}
I_{r}=\sum_{k=1}^{3}\lambda _{k}G_{r}(I_{s_{i_{k}}};H(\mathbf{p}_{t})),
\end{equation}
where $\mathbf{p}_{t}$ are the 2D landmarks of $F_{t}$. If any vertices of
the triangle are boundary points, we exclude them from the interpolation and
normalize the weights, $\lambda _{i}$, to sum to one.

A face view query is illustrated in Figs.~\ref{fig:interpolation}(b) and~\ref%
{fig:interpolation}(c). To improve the interpolation accuracy, we used a
horizontal flip to fill in views when the appearance map is one-sided with
respect to the yaw dimension, and generate artificial views using $G_{r}$
when the appearance map is too sparse.

\subsection{Face inpainting}

\label{subsec:inpainting}

A common challenge in face swapping is partial occlusions of the input face
image, $F_{s}$. Such parts of $F_{s}$ might be missing and thus cannot be
rendered on the target face, $F_{t}$. Nirkin et al.~\cite{nirkin2018face}
used the segmentation of $F_{s}$ and $F_{t}$ to remove occluded parts and
only rendered (and swapped) the face parts in the intersection of the masks.
Pixels in $F_{t}$ were used to render the facial regions occluded in $%
F_{s} $.

For significant occlusions and when the textures of $F_{s}$ and $F_{t}$
differ notably, the resulting face swap might have noticeable artifacts. To
overcome this, we propose to apply the face inpainting generator, $G_{c}$,
shown in Fig.~\ref{fig:system}(b), to render $F_{s}$ such
that the resulting face rendering, $I_c$, will cover the entire
segmentation mask, $S_{t}$ (of $F_{t}$), thus, resolving any occlusion issues.

This is a particular inpainting task in which the expression of the
occluded face is given, and the inpainted object is symmetric. These
significant cues were not utilized in our previous work \cite%
{nirkin2019fsgan}. To this end, in addition to the reenactment image and
segmentation mask, we also feed the inpainting generator with the target
face landmarks and the horizontal flip of the reenactment image.

Given an image, $I\in R^{3\times H\times W}$, landmarks
points, $p\in R^{98\times 2}$, and a segmentation mask, $S\in
R^{3\times H\times W}$, let $\tilde{I}$ be the input image
with its background removed, $\tilde{I}^{\prime }$ the same image
horizontally flipped, $H(p)\in R^{98\times H\times W}$ the heatmap
representation of $p$, and $\tilde{S}\in
R^{H\times W}$ a binary face mask derived from $S$. We
define the inpainting generator
\begin{equation}
G_{c}:\left\{\tilde{I},\tilde{I}^{\prime },H(p),\tilde{S}\right\} \rightarrow \mathbb{R}^{3 \times H \times W}.
\end{equation}

\subsubsection{Training}

\label{subsubsec:inpainting_training}

Given the reenactment result, $I_{r}$, its corresponding
segmentation, $S_{r}$, the target image with its background removed, %
$\tilde{I}_{t}$, and its corresponding face landmarks, $p_{t}$%
, we augment $S_{r}$ by simulating common face occlusions
due to hair, by randomly removing ellipse-shaped parts, in various sizes and
aspect ratios from the border of $S_{r}$. Let $\tilde{I}_{r}$
be $I_{r}$ with its background removed using the
augmented version of $S_{r}$, and $I_{c}$ the result of
applying $G_{c}$ to $\tilde{I}_{r}$ with the additional
input as described in the previous section. We define the inpainting
generator loss
\begin{equation}
\mathcal{L}(G_{c})=\lambda _{rec}\mathcal{L}_{rec}(I_{c},\tilde{I}%
_{t})+\lambda _{adv}\mathcal{L}_{adv},
\end{equation}%
where $L_{rec}$ and $L_{adv}$ are the
reconstruction and adversarial losses as in Section \ref%
{subsubsec:Reconstruction-Loss} and \ref{subsubsec:Adversarial-Loss},
respectively.

\subsection{Face blending}

\label{subsec:Blending}

The final phase of our face swapping scheme is the blending of the completed
face, $F_{c}$, with the target face, $F_{t}$, as depicted in Fig. \ref%
{fig:system}(c). The blending accounts for different skin tones and lighting
conditions, and is inspired by previous works that applied the Poisson
equation to inpainting \cite{yeh2017semantic} and blending \cite{wu2017gp}.
For that, we propose a novel Poisson blending loss.

Let $I_{t}$ be the target image, $I_{r}^{t}$ the image of the reenacted face
transferred onto the target image, and $S_{t}$ the segmentation mask marking
the transferred pixels. Following Perez et al.~\cite{perez2003poisson} we define the
Poisson optimization%
\begin{multline}
P(I_{t};I_{r}^{t};S_{t}))=\arg \min_{f}\Vert \nabla f-\nabla I_{r}^{t}\Vert
_{2}^{2}\newline
\text{ }  \label{equ:Poisson} \\
s.t.\text{ }f(i,j)=I_{t}(i,j),\text{ }\forall \text{ }S_{t}(i,j)=0,
\end{multline}%
where $\nabla \left( \cdot \right) $ is the gradient operator. We combine
the Poisson optimization above with the perceptual loss, thus deriving the
Poisson blending loss $\mathcal{L}(G_{b})$
\begin{equation}
\mathcal{L}(G_{b})=\lambda _{rec}\mathcal{L}%
_{rec}(G_{b}(I_{t};I_{r}^{t};S_{t}),P(I_{t};I_{r}^{t};S_{t}))+\lambda _{adv}%
\mathcal{L}_{adv}.
\end{equation}

\subsection{Subject-specific finetuning}

\label{subsec:finetuning}

The face reenactment generator, $G_r$, pretrained on a
large amount of subjects, can be finetuned to a specific subject using as a
little as 400 iterations and the same training configuration as in Section~%
\ref{sub:training_details}. After finetuning $G_r$, is used to
render a specific subject in any pose and expression. Finetuning allows to
improve the quality of the rendering and reproducing subtle face details
such as the appearance of the teeth. This formulation of our method is
quantitatively explored in the Experiments results section. Unless
explicitly stated, we do not finetune the reenactment generator in our
experiments.

\subsection{Post-processing}

\label{subsec:post_processing}

The raw result of our face swapping pipeline, which was the final result in our previous method, exhibits a slight
saturation relative to the original target image, therefore directly pasting
the generated image into the full target image might result in noticeable
bounding box borders. Hence, we apply soft erosion to the target
segmentation mask, $S_{t}$, resulting in a seamless transition from
"face" to "background". The final cropped face swapping result, $%
I_{b}^{^{\prime }}$, is then given by
\begin{equation*}
I_{b}^{^{\prime }}=I_{b}\cdot S_{t}+I_{t}\cdot (1-S_{t}).
\end{equation*}%
$I_{b}^{^{\prime }}$ is then seamlessly integrated into the full
target image by resizing it to the appropriate size and copying it to the
location of the detected bounding box.

\section{Datasets and training}

\label{sec:datasets_and_processing}

\subsection{Datasets and processing}

We used the video sequences of the IJB-C dataset~\cite{maze2018iarpa} to
train the generator, $G_{r}$, from which we automatically extracted the frames
containing particular subjects. IJB-C contains 11,000 face videos of which we
used 5,500 that were in high definition. Similar to the frame pruning
approach described in Section~\ref{subsec:FaceViewInterpolation}, we prune the
face views that are too close together as well as motion-blurred frames.

We apply the segmentation CNN $G_{s}$, to the frames and prune the frames
for which less than 15\% of the pixels in the face bounding box were
classified as face pixels. Dlib's face verification\footnote{%
Available: \url{http://dlib.net/}} was applied to group frames according to
the depicted subjects' identity, and limit the number of frames per subject
to 100, by choosing the frames whose 2D landmarks have the maximal variance.
In each training iteration, we choose the frames, $I_{s}$ and $I_{t}$, from
two randomly chosen subjects.

The VGG-19 CNNs were trained using perceptual losses on the VGGFace2 dataset~%
\cite{cao2018vggface2} for face recognition and the CelebA~\cite%
{liu2018large} dataset for face attribute classification. The VGGFace2
dataset contains 3.3M images of 9,131 identities, whereas CelebA
contains 202,599 images. The segmentation CNN, $G_{s}$, was trained using
the dataset from~\cite{nirkin2018face}, consisting of ${\sim }10K$ face
images labeled with face segmentations, and the LFW Parts Labels Dataset~%
\cite{kae2013augmenting} with ${\sim }3K$ images labeled for both face and
hair segmentations, from which we removed the neck part using facial
landmarks.

We used additional 1K images and corresponding hair segmentations from the
Figaro dataset~\cite{svanera2016figaro}. Finally, FaceForensics++~\cite%
{roessler2019faceforensics++} provides 1000 videos, from which they
generated 1000 synthetic videos on random pairs using DeepFakes~\cite%
{DeepFakes} and Face2Face~\cite{thies2016face2face}.

\subsection{Training details}

\label{sub:training_details}

The proposed generators were trained from scratch, where the weights were
initialized randomly using a normal distribution. We used the Adam optimizer~%
\cite{kingma2014adam} ($\beta _{1}=0.5,\beta _{2}=0.999$) and a learning
rate of $0.0002$ that was reduced by half every ten epochs. The following
parameters were used for all the generators: $\lambda _{perc}=1,\lambda
_{pixel}=0.1,\lambda _{adv}=0.001,\lambda _{seg}=0.1,\lambda
_{rec}=1,\lambda _{stepwise}=1$, where $\lambda _{reenactment}$ is linearly
increased from 0 to 1 during the training process. All of our networks were
trained using eight NVIDIA Tesla V100 GPUs, where the training of the $G_{s}$
generator required six hours to converge, while the rest of the networks
converged in two days. All our networks except for $G_{s}$, were trained
using a progressive multi-scale approach, starting with a resolution of 128$%
\times $128 up to 256$\times $256. The reenactment inference rate is ${\sim }%
30$fps, and ${\sim }10$fps for face swapping, both using a single reenactment iteration on a single NVIDIA Tesla V100 GPU.

\section{Experimental results}

We performed extensive qualitative and quantitative trials to experimentally
verify the proposed scheme and compare our method to contemporary face
swapping methods: DeepFakes~\cite{DeepFakes} and Nirkin et al.~\cite%
{nirkin2018face}, and the Face2Face reenactment scheme~\cite%
{thies2016face2face}. We conduct all our experiment using FaceForensics++~%
\cite{roessler2019faceforensics++} videos, running our method on the same
pairs they used. Moreover, we provide an ablation study to exemplify the
importance of each component in our pipeline.

\subsection{Quantitative experiments metrics}

\label{subsec:quant_experiments_metrics}

\begin{figure*}[t]
\centering
\includegraphics[width=1.0\linewidth]{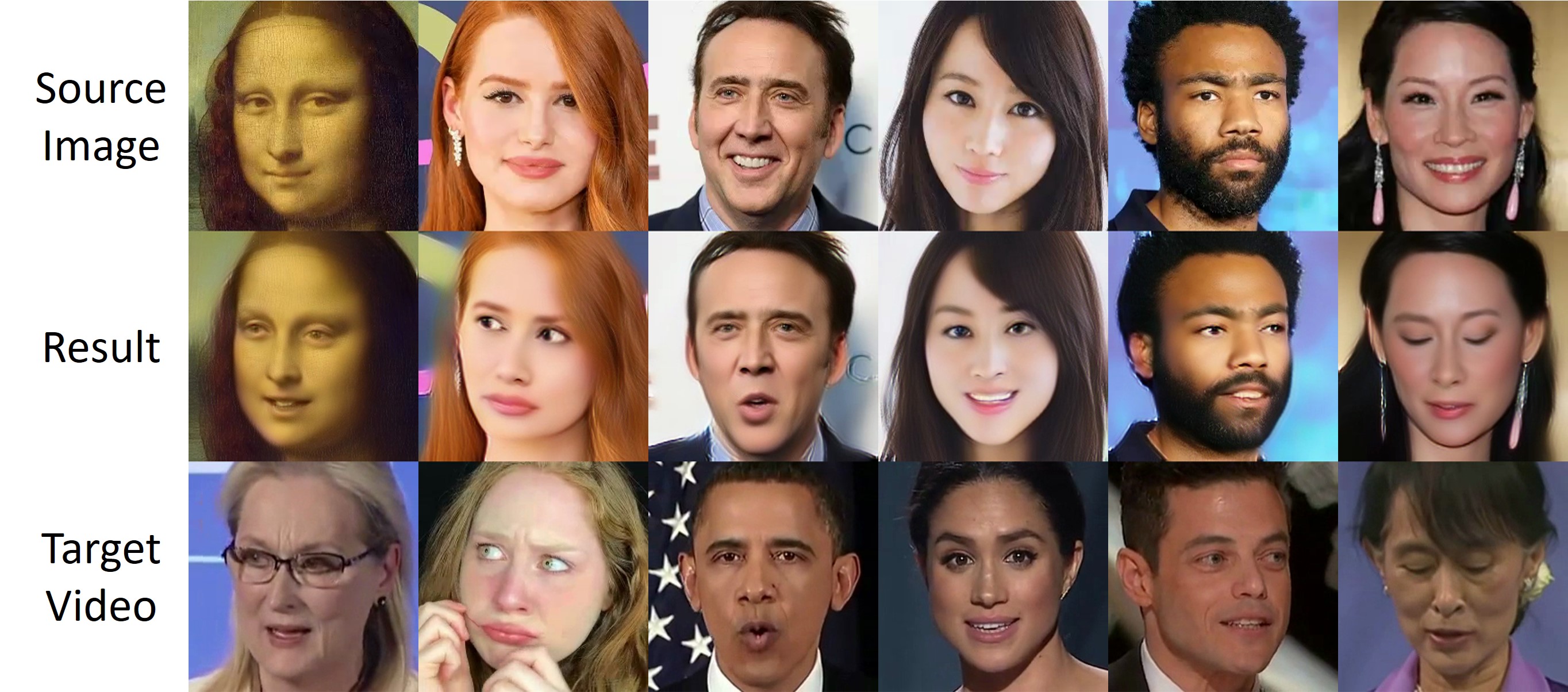}
\caption{ Qualitative face reenactment results. Row 1: The source face for
reenactment. Row 2: Our reenactment results (without background removal).
Row 3: The target face from which to transfer the pose and expression.
\protect\vspace{-4mm} }
\label{fig:face_reenactment_qualitative}
\end{figure*}

We applied multiple metrics and corresponding quantitative tests to
experimentally evaluate our face swapping scheme in general, and the
identity preservation, in particular, while retaining the same pose and
expression as the target subject. For that, we used the following metrics:

\vspace{2mm}\noindent {\textbf{Identity verification (id)}.} To measure the
identity preservation of our method, we first extract face embeddings using Arcface~\cite{deng2018arcface}, and then compare them using cosine similarity:
\begin{equation}
\text{similarity}=\langle \frac{a}{\Vert a \Vert _{2}}, \frac{b}{\Vert b \Vert _{2}} \rangle
\end{equation}%
where $a$ and $b$ are the face embeddings.
This face verification network is a different network than the one used for face-specific perceptual loss, and it was trained on a different dataset.

\vspace{2mm}\noindent {\textbf{Fr\'{e}chet Inception Distance (FID)}.} To
asses the visual quality of our generated images we adopt the FID metric~%
\cite{heusel2017gans} which is more stable than previous quality metrics
such as the structural similarity index method (SSIM) or the Inception score~%
\cite{salimans2016improved}.

\vspace{2mm}\noindent {\textbf{$\mathcal{L}_{1}$ distance}.} A quality
measurement that focuses on the low frequencies of the images, validating
the color preservation in the generated image. For this metric we compare
images with color values in the range $[0,1]$.

\vspace{2mm}\noindent {\textbf{Euler angles}.} To compare the face poses we
calculate the Euclidean distance between the Euler angles of the generated
face and those of the target face, using the method from~\cite{ruiz2018fine}%
. This distance is quantified in degrees.

\vspace{2mm}\noindent {\textbf{Face landmarks}.} Similarly to the Euler
angles metric, we measure the accuracy of the facial expression and shape by
calculating the Euclidean distance between the face landmarks of the
generated face and the target face. This distance is measured in pixels.

\vspace{2mm}\noindent {\textbf{Facial expression comparison (FEC)}. }

To further validate the accuracy of the generated facial expressions, we add the FEC metric. This is an alternative to the expression classification evaluation used for the same purpose by Chang et al.~\cite{chang2018expnet}, that used a limited number of expressions. The FEC metric is directly trained to compare facial expressions and mimics the way humans perceive expressions.

To this end, we train an expression embedding network using the FEC dataset~\cite%
{vemulapalli2019compact}, consisting of 156K images and 500K human-annotated
triplets, such that each triplet contains a pair of expressions that are
mutually more similar to the third expression. Following Vemulapalli et al.~\cite%
{vemulapalli2019compact} we use the triplet loss~\cite{schroff2015facenet}
to train an embedding network to compute a 16-dimensional expression
embedding.

\subsection{Face reenactment results}

\label{sub:face_reenactment_results}

\subsubsection{Qualitative face reenactment results}

\label{subsub:qualitative_face_reenactment_results}

We start by depicting qualitative face reenactment results in Fig.~\ref%
{fig:face_reenactment_qualitative}. We show raw face reenactment results, of
varying ethnicities, poses, and expressions without background removal.
In the left-most column we demonstrate reenacting a face
that differs significantly from the training set, exemplifying that our
method performs well when applied to faces from completely
different domains. The second column from the left demonstrates our method's
ability to cope with rare facial expressions that are not present in the
training set, and extreme iris position differences.

To exemplify the importance of the iterative reenactment method, we illustrate
in Fig.~\ref{fig:reenactment_limitations} the reenactment of the same
subject for both small and large angular differences. For large angular
differences, the identity and texture are better preserved using multiple
iterations than by using a single one.

\subsubsection{Pose only reenactment}

\label{subsub:pose_only_reenactment}

\begin{figure*}[t]
\centering
\includegraphics[width=1.0\linewidth]{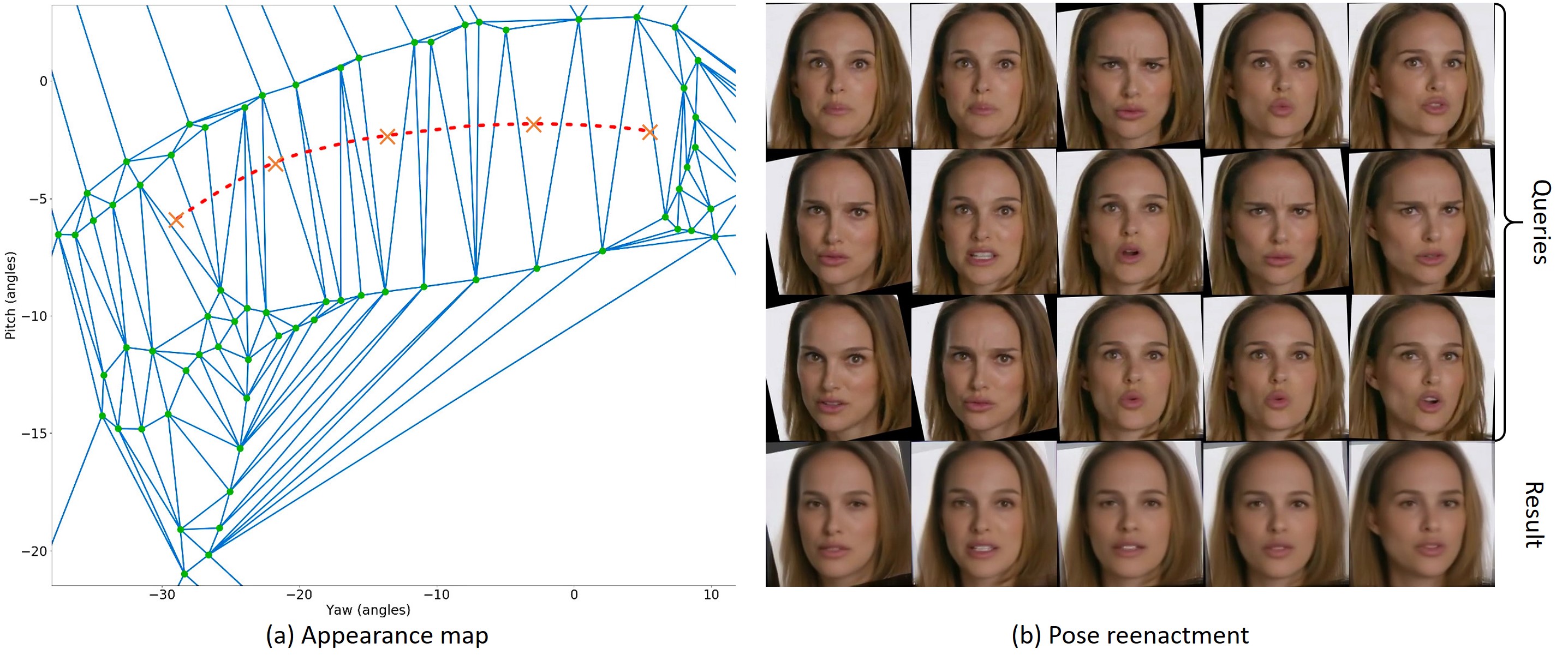}
\caption{Pose reenactment. (a) Example appearance map of an input
subject (Natalie Portman). The red dashes represent the interpolation path
and the orange Xs mark the pose of the views presented on the right. (b) The
first three rows are input queries from the appearance map. The reenactment
result is displayed in the last row.}
\label{fig:pose_reenactment}
\end{figure*}

Using the reenactment generator, $G_r$, and the landmarks
transformer, $T$, we can reenact a source face without a target
face, using only Euler angles to guide the pose. For that, we apply the face
view interpolation in Section~\ref{subsec:FaceViewInterpolation}, where the
Euler angles were varied smoothly by a user using a mouse or keyboard. For
each Euler angle, the appearance map (as in Section \ref%
{subsec:FaceViewInterpolation}) of the input subject is queried for three
face views, which are then rotated to the same tilt angle as the current
Euler angle. The views are then interpolated the same way as in Section~\ref%
{subsec:FaceViewInterpolation}. Such a reenactment is shown in Fig.~\ref%
{fig:pose_reenactment}.

\subsubsection{Expression-only reenactment comparison}

\label{subsub:expression_only_reenactment}

\begin{figure}[t]
\includegraphics[width=1.0\linewidth]{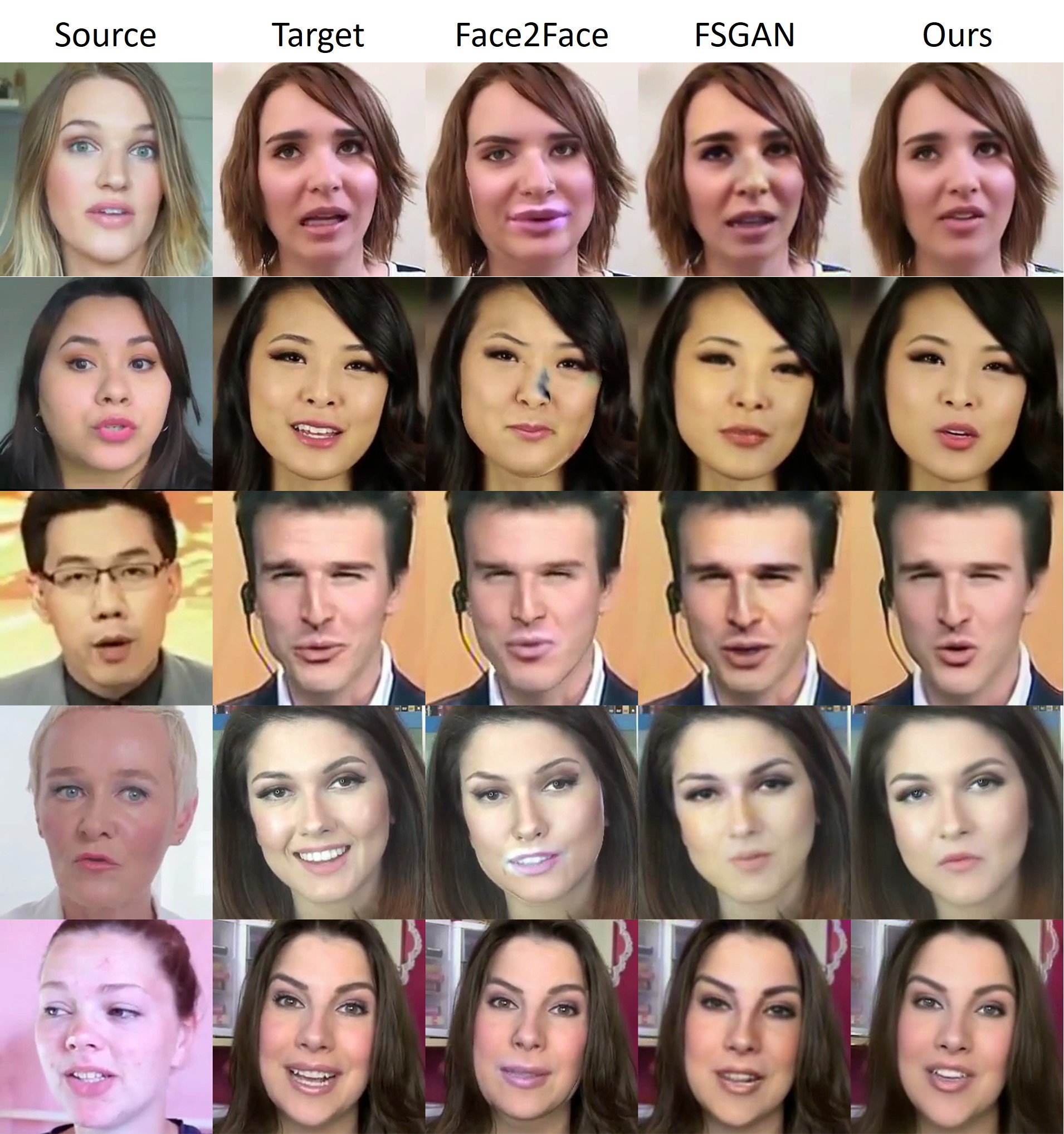}
\caption{\emph{Qualitative expression-only reenactment comparison.}
The expression of the face in the left column is transferred to the face in
the second column using Face2Face~\protect\cite{thies2016face2face} (third
column), FSGAN~\protect\cite{FaceSwap} (4th column), and our method (last
column).} \protect\vspace{-4mm}
\label{fig:face2face}\centering
\end{figure}

\begin{figure}[tbp]
\centering
\includegraphics[width=1.0%
\linewidth]{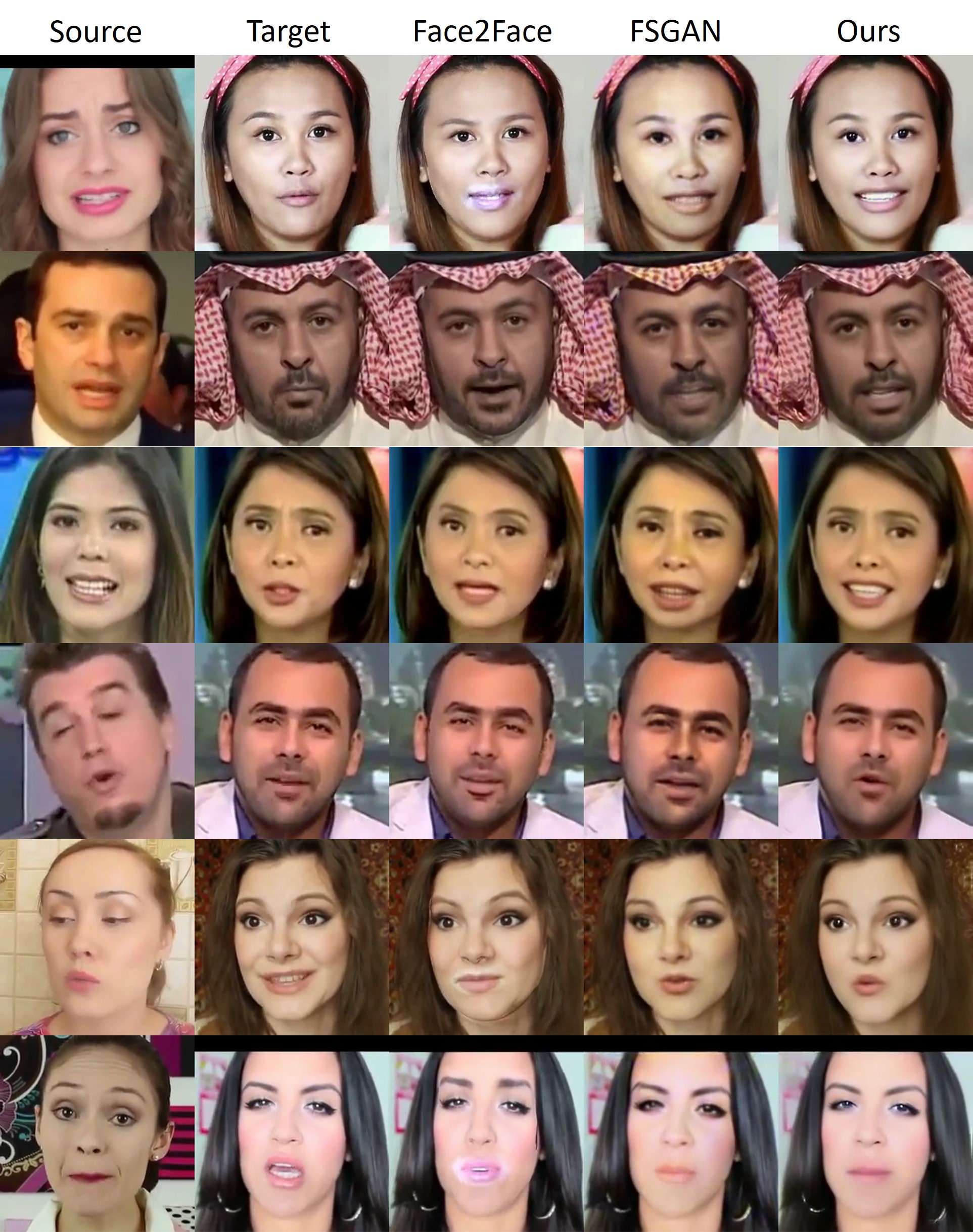}
\caption{Additional qualitative expression-only reenactment
comparison on FaceForensics++~\protect\cite{roessler2019faceforensics++}.}
\label{fig:additional_face2face}
\end{figure}

We compare our method to the contemporary Face2Face~\cite%
{thies2016face2face} and FSGAN~\cite{nirkin2019fsgan} methods in the
expression-only reenactment problem, and the results are shown in Fig.~\ref%
{fig:face2face} and Fig.~\ref{fig:additional_face2face}. For this problem we use a slightly different formulation:
Given a pair of faces, $F_{s}\in I_{s}$ and $F_{t}\in I_{t},$ we aim to
transfer the expression from $I_{s}$ to $I_{t}$. Hence, we modify the
corresponding 2D landmarks of $F_{t}$ by swapping in the mouth points of the
2D landmarks of $F_{s},$ similar to the generation of the intermediate
landmarks in Section~\ref{subsec:reenactment}. The reenactment result is
then given by $G_{r}(I_{t};H(\hat{p}_{t}))$ where $\hat{p}_{t}$ are the
modified landmarks.

\subsection{Face inpainting ablation results}

\label{subsec:inpainting_results}

\begin{table*}[tbp]
\centering{\
\resizebox{1.0\linewidth}{!}{
 \begin{tabular}{lcccccc}
 \toprule
 Method & FID $\downarrow$ & id $\uparrow$ & $\mathcal{L}_1$ $\downarrow$ & euler $\downarrow$ & landmarks $\downarrow$ & FEC $\downarrow$ \\ [0.5ex]
 \hline
 FSGAN~\cite{nirkin2019fsgan} & 2.64 & 0.36 $\pm$ 0.07 & 0.46 $\pm$ 0.01 & 1.77 $\pm$ 0.92       & 34.8 $\pm$ 15.0 & 0.15 $\pm$ 0.09 \\
 $G_c$ without landmarks      & 0.74 & \textbf{0.64 $\pm$ 0.11} & 0.31 $\pm$ 0.02 & 1.20 $\pm$ 0.75       & 19.4 $\pm$ 7.5  & \textbf{0.11 $\pm$ 0.09} \\
 $G_c$ with landmarks & \textbf{0.62} & \textbf{0.64 $\pm$ 0.10} & \textbf{0.16 $\pm$ 0.01} & {\bf 1.04 $\pm$ 0.52} & {\bf 17.9 $\pm$ 3.8} & 0.12 $\pm$ 0.07 \\
 \bottomrule
 \end{tabular}
}
}
\caption{\emph{Inpainting ablation.} We compare our previous inpainting results ~\protect\cite{nirkin2019fsgan} with two variations of the proposed face inpainting method, the first (middle row) without using
face landmarks and the seconds (bottom row) with face landmarks alongside a horizontally flipped replica of the source image.}
\label{tab:inpainting_ablation}
\end{table*}

\begin{figure*}[t]
\centering
\includegraphics[width=1.0\linewidth]{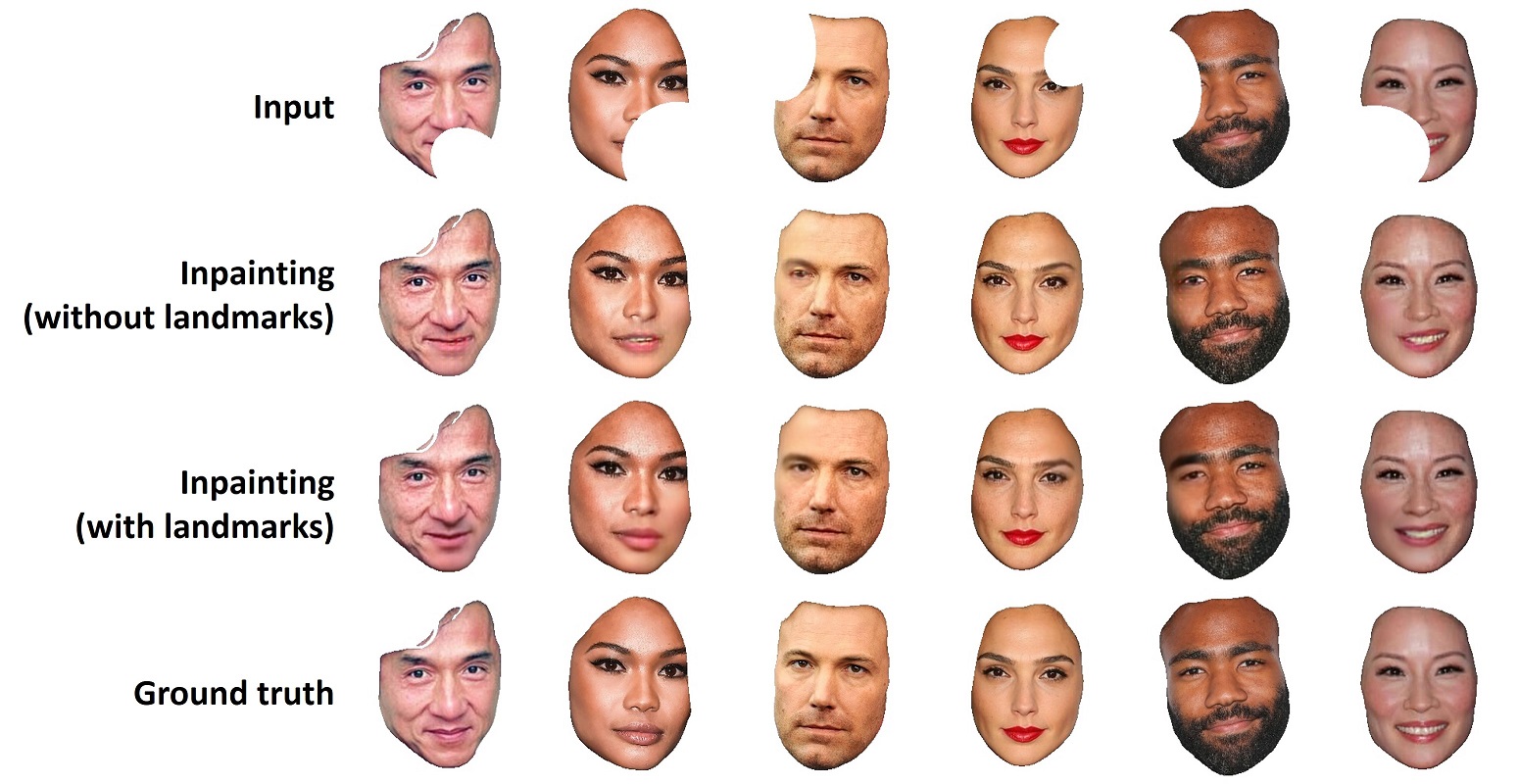}
\caption{\emph{\ Face inpainting qualitative results.} Top row:
Input real images with random ellipses removed from the faces. Seconds row:
Inpainting results without facial landmarks. Third row: Inpainting results
with facial landmarks. Bottom row: The full input image.}
\label{fig:inpainting_qualitative}
\end{figure*}

To exemplify the effectiveness of our proposed improvements of $G_c$, 
we perform an ablation study. In
this study we remove random ellipses from faces in the first 100 frames of
the first 500 videos in FaceForensics++, similar to Section~\ref%
{subsubsec:inpainting_training}, inpaint the face using the different
inpainting methods and compare the inpainting results to the original faces.

Qualitative examples of the ablation study are shown in Fig.~\ref%
{fig:inpainting_qualitative}. It follows that our \textit{new} FSGAN
architecture for $G_{c}$ produces higher quality results than the
\textit{previous} FSGAN\cite{nirkin2019fsgan}. Moreover, the FSGAN
formulation that receives the face landmarks as an additional input better
reconstructs the facial expressions and shapes of the input faces.
Quantitative results are shown in Table~\ref{tab:inpainting_ablation} using
the metrics detailed in Section~\ref{subsec:quant_experiments_metrics}.

As apparent from the table, both new FSGAN formulations significantly outperform the
previous FSGAN \cite{nirkin2019fsgan} in all metrics. Our $G_{c}$
formulation that also receives the face landmarks and the horizontal flip of
the image, outperforms the variant without those additions on all metrics,
but for the id metric, for which both variants achieve similar scores.

\subsection{Face swapping results}

\label{sub:face_swapping_results}

\subsubsection{Qualitative face swapping results}

\label{subsub:qualitative_face_swapping_results}

\begin{figure*}[t]
\centering
\includegraphics[width=1.0\linewidth]{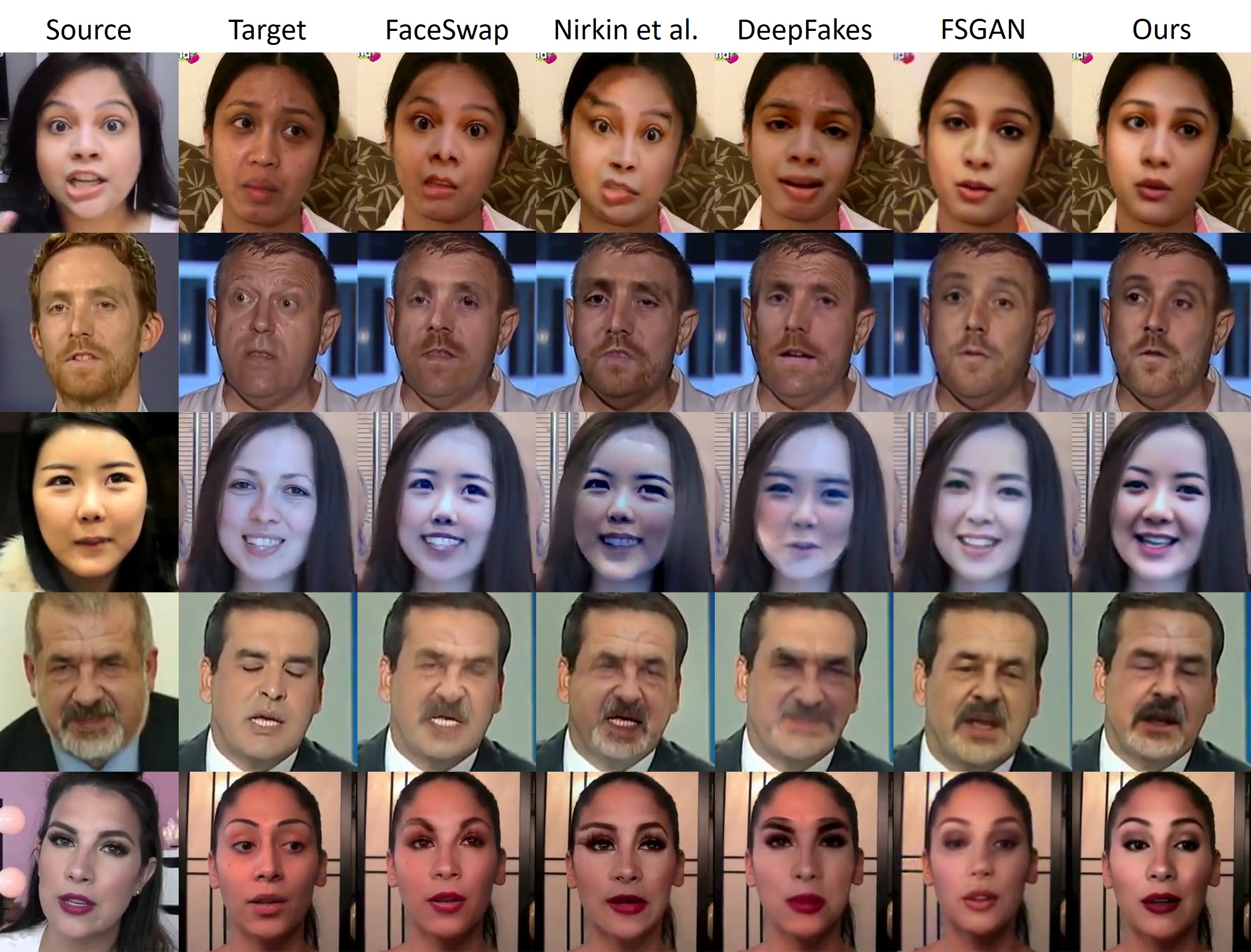}
\caption{ \emph{Qualitative face swapping results on~\protect\cite%
{roessler2019faceforensics++}.} The 3rd and 4th columns shows the
results of the 3DMM-based methods FaceSwap~\protect\cite{FaceSwap} and
Nirkin et al.~\protect\cite{nirkin2018face}, the 5th column shows DeepFakes~%
\protect\cite{DeepFakes} results, a direct mapping method, which was trained
specifically on each of those videos. The 6th column shows FSGAN~%
\protect\cite{nirkin2019fsgan} results and the last column shows the results
of our method, both on images of faces they have not seen before. \protect%
\vspace{-4mm} }
\label{fig:face_swap_qualitative}
\end{figure*}

\begin{figure*}[tbp]
\centering
\includegraphics[width=%
\linewidth]{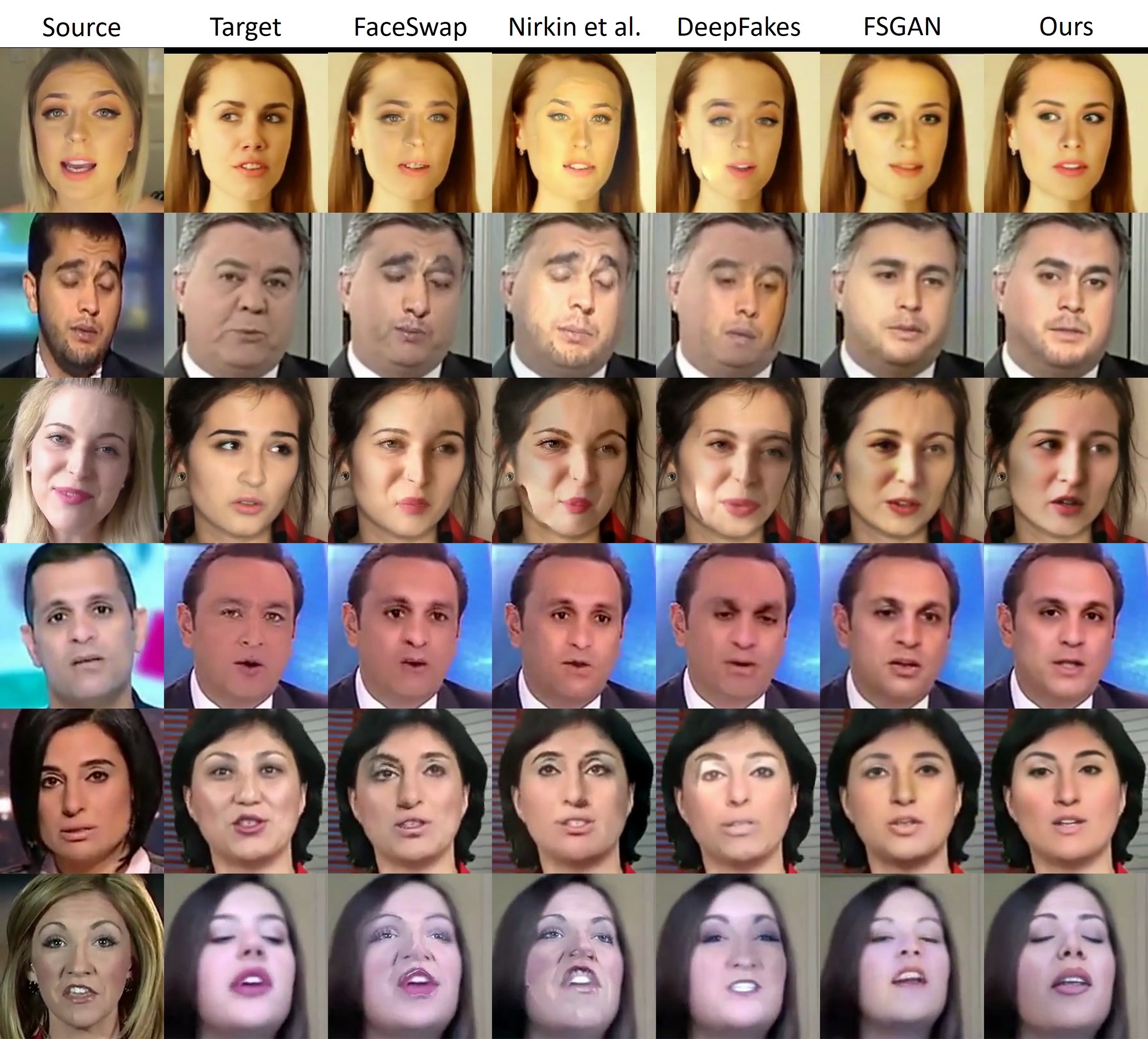}
\caption{ Additional qualitative face swapping comparison to previous
methods on FaceForensics++~\protect\cite{roessler2019faceforensics++}.}
\label{fig:additional_comparison}
\end{figure*}

Qualitative results of our face swapping method applied to the
FaceForensics++ videos \emph{without training our model on any of them }are
shown in Fig.~\ref{fig:face_swap_qualitative}. We purposefully chose
examples showing different poses and expressions, face shapes, and hair
occlusions. Nirkin et al.~\cite{nirkin2018face} is an image-to-image face
swapping method and therefore to be applied to videos it requires a selection strategy for the source frames. Thus, to be fair in our comparison, for each frame in the
target video, we select the source frame having the most similar pose, as face swapping becomes more difficult the larger the pose discrepancy.
To compare FSGAN in a video-to-video scenario such as this, we use our face
view interpolation, as described in Section~\ref%
{subsec:FaceViewInterpolation}. Additional qualitative comparison is presented in Fig.~\ref{fig:additional_comparison}.

\subsubsection{Quantitative face swapping results}

\label{subsubsec:face_swapping_quant}

\begin{table*}[t]
\centering{\
\resizebox{1.0\linewidth}{!}{
 \begin{tabular}{lcccccc}
 \toprule
 Method & FID $\downarrow$ & id $\uparrow$ & $\mathcal{L}_1$ $\downarrow$ & euler $\downarrow$ & landmarks $\downarrow$ & FEC $\downarrow$ \\ [0.5ex]
 \hline
 Nirkin et al.~\cite{nirkin2018face} & 0.67 & 0.45 $\pm$ 0.06 & 0.38 $\pm$ 0.03 & 3.28 $\pm$ 1.34 & 42.1 $\pm$ 9.19 & 0.37 $\pm$ 0.18 \\
 FaceSwap~\cite{FaceSwap} & \textbf{0.37} & 0.38 $\pm$ 0.05 & 0.44 $\pm$ 0.02     & 2.43 $\pm$ 0.85 & 36.9 $\pm$ 7.35 & 0.34 $\pm$ 0.17 \\
 DeepFakes~\cite{DeepFakes} & 0.38 & 0.45 $\pm$ 0.04 & 0.44 $\pm$ 0.02   & 4.05 $\pm$ 1.13 & 53.1 $\pm$ 8.32 & 0.45 $\pm$ 0.14 \\
 FSGAN~\cite{nirkin2019fsgan} & 0.46 & 0.36 $\pm$ 0.06 & 0.25 $\pm$ 0.01 & 2.37 $\pm$ 1.29 & 30.8 $\pm$ 11.2 & 0.34 $\pm$ 0.14 \\
 FaceShifter~\cite{li2019faceshifter} & 0.40 & \textbf{0.60 $\pm$ 0.03} & 0.42 $\pm$ 0.11 & 2.33 $\pm$ 0.8 & 45.51 $\pm$ 11.36 & 0.29 $\pm$ 0.11 \\
 Ours (fine-tuned) & 0.50 & 0.44 $\pm$ 0.04  & 0.22 $\pm$ 0.01          & 2.40 $\pm$ 0.77 & 28.7 $\pm$ 5.76 & 0.30 $\pm$ 0.11 \\
 Ours              & 0.53 & 0.37 $\pm$ 0.05 & \textbf{0.21 $\pm$ 0.0}  & \textbf{2.07 $\pm$ 0.68} & \textbf{21.6 $\pm$ 5.25} & \textbf{0.28 $\pm$ 0.10} \\
 \bottomrule
 \end{tabular}
}
}
\caption{Quantitative face swapping comparison. We compare to
previous face swapping methods and to a fine-tuned formulation of our method
trained using the FaceForensics++~\protect\cite{roessler2019faceforensics++}.}
\label{tab:face_swapping_quant}
\end{table*}

We also applied quantitative tests to compare our entire face
swapping pipeline with previous face swapping schemes. For that, we used the
same videos as in Section~\ref{subsec:inpainting_results}, and the metrics
detailed in Section~\ref{subsec:quant_experiments_metrics}. We compute the
mean and standard deviation, averaged across the videos, for all the metrics
except for the FID metric that is computed globally. The metrics $L_{1}$, 
Euler, landmarks, and FEC, for each frame, are compared to the
corresponding frame in the target video, to test color, shape, pose, and
expression preservation of the target face. 
The id metric for each frame, is compared against random source frames to quantify how well the identity is retained. Finally, the FID metric is used to compare the texture distributions of the source and generated faces.

We compare with five previous face swapping methods: Nirkin et al.~%
\cite{nirkin2018face}, FaceSwap~\cite{FaceSwap}, DeepFakes~\cite{DeepFakes}, FSGAN~\cite{nirkin2019fsgan}, and FaceShifter~\cite{li2019faceshifter}. The first column of Table~\ref%
{tab:face_swapping_quant} shows that the FaceSwap approach achieves the best
FID score as it directly warps the texture from the source image. Our method
achieves a slightly better FID score as explained in Section~\ref%
{subsec:limitations}. Our fine-tuned version attains an improved FID score
as it is further trained using the source images. 
The second column shows that FaceShifter, a recent state-of--of-the--the-art method, best preserves the identity of the source subject. Our method achieves a lower identity score, yet retains color, pose, and expression significantly better than previous methods, as shown the four rightmost metrics in Table~\ref{tab:face_swapping_quant}. As can be seen from the results, there is a trade-off between how much a method can preserve the source identity and faithfully maintain other important attributes such as: color, pose, and expression.

\subsection{Ablation study on pipeline components}

\begin{figure*}[tbh]
\centering
\includegraphics[width=1.0\linewidth]{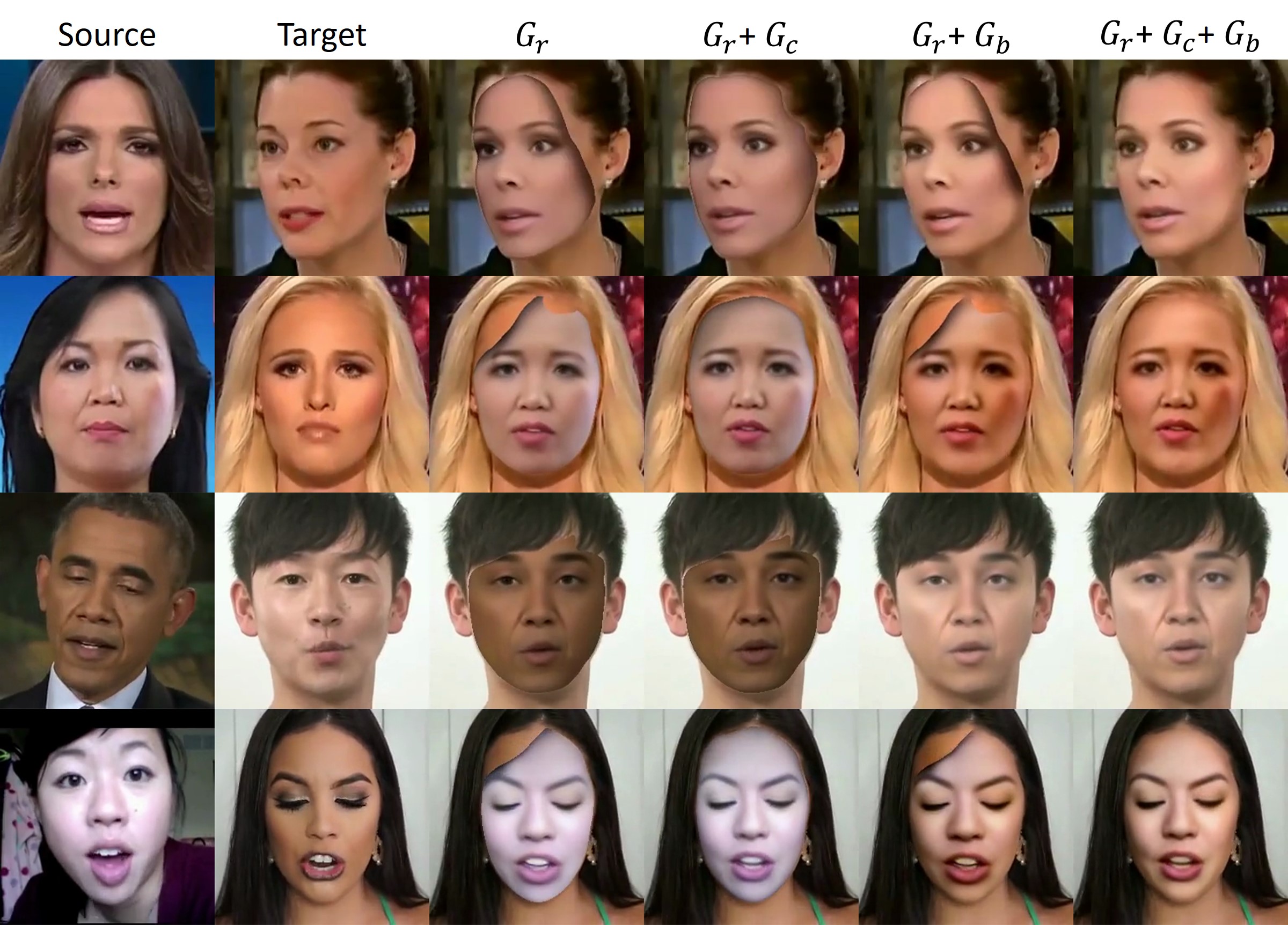}\vspace{%
-2mm} \vspace{-4mm}
\caption{Ablation study on the pipeline components. As shown in columns 3
and 5, without the completion network, $G_{c}$, the transferred face won't
cover the entire target face leaving obvious artifacts. In columns 3 and 4,
we can see that without the blending network, $G_{b}$, the skin color and
lighting conditions of the transferred face are inconsistent with its new
context.}
\label{fig:ablation_qualitative}
\end{figure*}

\begin{table*}[tbh]
\centering{\
\resizebox{1.0\linewidth}{!}{
\begin{tabular}{lcccccc}
\toprule
Method & FID $\downarrow$ & id $\uparrow$ & $\mathcal{L}_1$ $\downarrow$ & euler $\downarrow$ & landmarks $\downarrow$ & FEC $\downarrow$ \\ [0.5ex]
\hline
FSGAN $(G_r)$        & 0.72          & 0.44 $\pm$ 0.06 & 0.26 $\pm$ 0.01 & 2.93 $\pm$ 1.11 & 32.9 $\pm$ 8.46 & 0.36 $\pm$ 0.13 \\
Ours $(G_r)$         & 0.71          & \textbf{0.45} $\pm$ 0.07 & 0.22 $\pm$ 0.01          & 2.93 $\pm$ 0.91 & 31.8 $\pm$ 5.66          & 0.30 $\pm$ 0.11 \\
Ours $(G_r+G_c)$     & 0.67          & 0.41 $\pm$ 0.06 & 0.21 $\pm$ 0.01          & 2.57 $\pm$ 0.83 & 23.2 $\pm$ 4.95          & 0.29 $\pm$ 0.10 \\
Ours $(G_r+G_b)$     & 0.54          & 0.41 $\pm$ 0.06 & \textbf{0.20 $\pm$ 0.01} & 2.40 $\pm$ 0.79 & 28.0 $\pm$ 5.17          & 0.29 $\pm$ 0.11 \\
Ours $(G_r+G_c+G_b)$ & \textbf{0.53} & 0.37 $\pm$ 0.06 & 0.21 $\pm$ 0.0  & \textbf{2.07 $\pm$ 0.68} & \textbf{21.6 $\pm$ 5.25} & \textbf{0.28 $\pm$ 0.10} \\
\bottomrule
\end{tabular}
} }
\caption{Quantitative face swapping results on FaceForensics++~\protect\cite%
{roessler2019faceforensics++} for our ablation study on the components of
FSGAN.}
\label{tab:ablation_quant}
\end{table*}

To demonstrate the necessity of each component in our pipeline, we conducted
an ablation study on four different configurations of our method: $G_{r}$
only, $G_{r}$ with $G_{c}$, $G_{r}$ with $G_{b}$, and the entire pipeline.
The segmentation network, $G_{s}$, is used in all configurations. Qualitative
examples of our ablation study are depicted in Fig.~\ref%
{fig:ablation_qualitative}.

The quantitative results of our ablation study are shown in Table ~\ref%
{tab:ablation_quant}. 
Comparing our $G_r$ to our previous reenactment generator (first row), demonstrates an improvement in all metrics, indicating that the landmarks method described in Section~\ref{subsubsec:landmarks_transformer} is superior to the one in FSGAN~\cite{nirkin2019fsgan}.
Our entire pipeline achieves the best FID
score as shown in the first column, indicating that the texture of the
source face is best preserved when none of the steps are omitted. The id
scores (in the second column) show that the source subject identity is
preserved across all pipeline networks. The $L_{1}$ scores (in
the third column) confirms that the appearance of the target face are best
reconstructed when the blending generator, $G_{b}$, is applied,
indicating the importance of the final blending step. Finally, as
demonstrated by the last three columns, the pose and expression of the
target face are best preserved by our entire pipeline.

\section{Conclusions and Future work}

\subsection{Limitations}

\label{subsec:limitations}

\begin{figure}[tbh]
\centering
\includegraphics[width=1.0\linewidth]{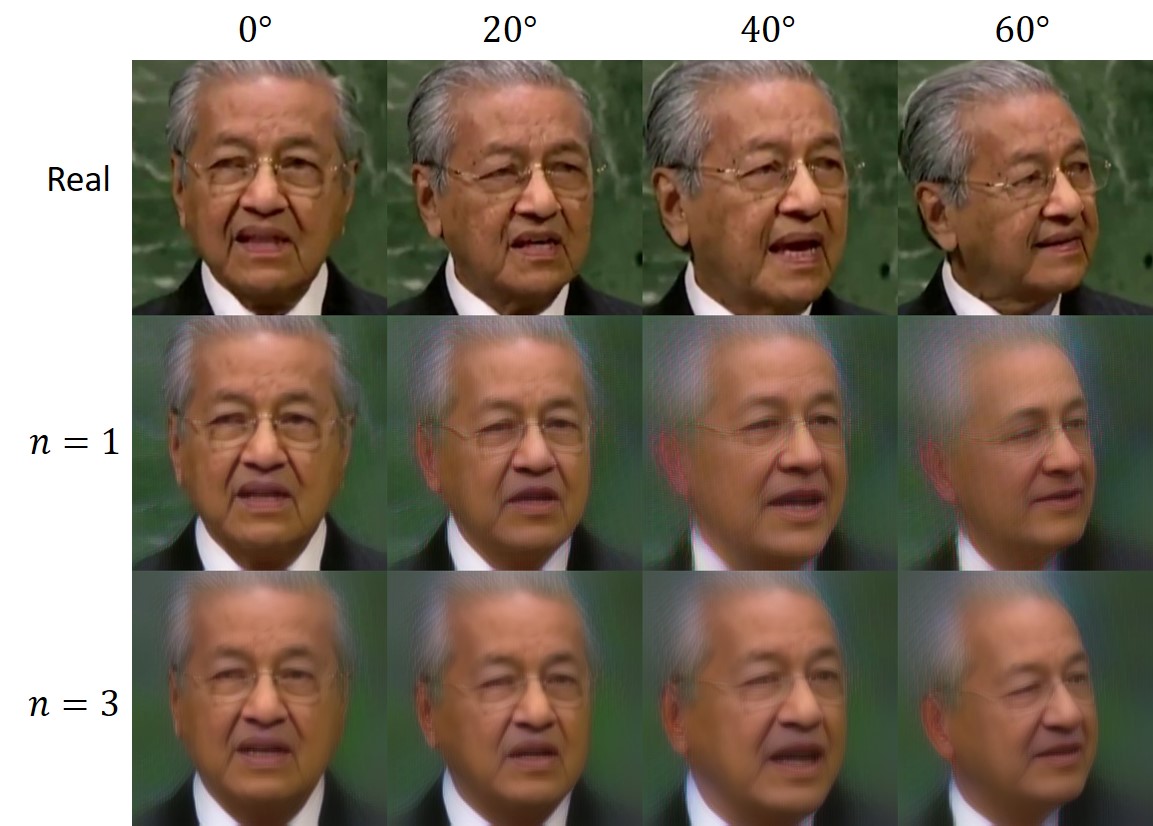}
\caption{ Reenactment limitations. Top left image transformed into each of
the images in Row 1 (using the same subject for clarity). Row 2: Reenactment
with one iteration. Row3: Three iterations. \protect\vspace{-4mm} }
\label{fig:reenactment_limitations}
\end{figure}

Figure~\ref{fig:reenactment_limitations} shows our reenactment results for
different facial yaw angles. Evidently, the larger the angular differences,
the more the identity and texture quality degrade. Moreover, applying too
many iterations of the face reenactment generator also blurs the texture. As
opposed to 3DMM based methods such as Face2Face~\cite{thies2016face2face},
which warp the texture directly from the image, our method is limited to the
image resolution of the training data. Another limitation arises from using
a sparse landmark tracking method that does not fully capture the complexity of the facial expressions.

In the video-to-video scenario, the proposed face view interpolation method assume an homogeneous image quality and facial expressions. If those assumptions do not hold, it might lead to artifacts in the interpolation results. Finally, there is a trade-off between how much our method can preserve the source identity while maintaining the attributes of the target subject such as: skin color, pose, and expression.

\subsection{Discussion}

Our method eliminates laborious, subject-specific data collection and model
training, making face swapping and reenactment accessible to non-experts. We
feel strongly that it is of \emph{paramount importance} to publish such
technologies, to drive the development of technical counter
measures for detecting such forgeries, as well as compel law makers to set
clear policies for addressing their implications. Suppressing the
publication of such methods would not stop their development, but rather
make them available to select a few and potentially blindside policy makers if
it is misused.

\subsection{Future work}

We have identified that the human factor is a major limitation of our
method. It is imposed as we rely on face and face landmarks detection, which
are trained using human labeling. Face landmarks in particular, are an
abstraction originally derived for humans, which filter-out significant
important information that could be otherwise utilized by the CNN. Future
research should avoid using methods that were trained using human labeling.
This would allow to achieve subpixel accurate face tracking and
segmentation, and improve image quality and resolution. Another direction would gain accuracy while reducing runtime by skipping face and landmark detection altogether, instead leveraging recent methods for direct, image to pose, face pose estimation~\cite{albiero2021img2pose} and facial expression deformation modeling~\cite{chang2019deep}. Previous reenactment
methods mainly focused on specific domains such as faces or bodies,
reenactment methods could potentially be generalized to be applied to
multiple domains.

Utilizing only one frame at a time is suboptimal as it does not make use of all the information available for the specific subject. Additionally, temporal information can help improve the tracking of the face pose and expression. Future methods should be able to take all this available information into consideration.

\bibliographystyle{IEEEtran}
\bibliography{egbib}

\vfill

\begin{IEEEbiography}[{\includegraphics[width=1in,height=1.25in,clip,keepaspectratio]{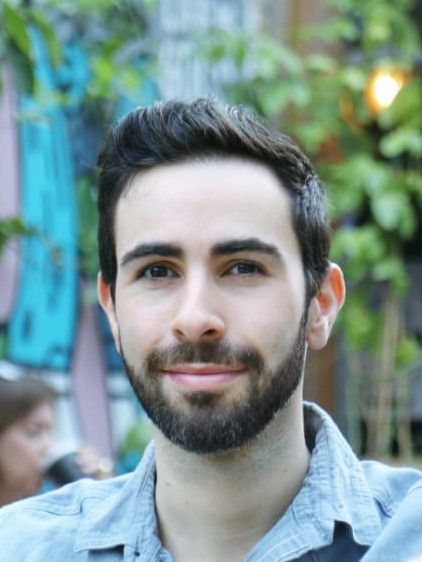}}]{Yuval Nirkin}
received his B.Sc. degree in computer engineering from the Technion Israel Institute of Technology, Haifa, in 2011, his M.Sc. in Computer Science from The Open University of Israel, Ra'anana, Israel, in 2017, and his Ph.D. in electrical engineering from Bar-Ilan University, Ramat Gan, Israel, in 2022. His research focuses on Deep Learning, Computer Vision, and Computer Graphics. He has been a reviewer in ECCV, ICCV, and CVPR, and was recognized as a high quality reviewer in ECCV'20.
\end{IEEEbiography}

\begin{IEEEbiography}[{\includegraphics[width=1in,height=1.25in,clip,keepaspectratio]{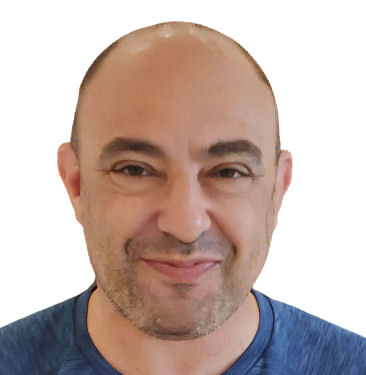}}]{Yosi Keller}
received the B.Sc. degree in electrical
engineering from the Technion Israel Institute
of Technology, Haifa, in 1994, and the M.Sc. and
Ph.D. degrees in electrical engineering from Tel
Aviv University, summa cum laude, in 1998 and
2003, respectively. From 2003 to 2006 he was a
Gibbs Assistant Professor with the Department
of Mathematics, Yale University, New Haven,
CT, USA. He is an Associate Professor at the
Faculty of Engineering in Bar Ilan University,
Ramat-Gan, Israel. His research interests are
in Computer Vision, Machine and Deep Learning, and Biometrics.
\end{IEEEbiography}

\begin{IEEEbiography}[{\includegraphics[width=1in,height=1.25in,clip,keepaspectratio]{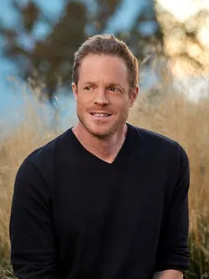}}]{Tal Hassner}
received his M.Sc. and Ph.D. degrees in applied mathematics and computer science from the Weizmann Institute of Science, Israel, in 2002 and 2006, respectively. In 2008 he joined the Department of Math. and Computer Science at The Open Univ. of Israel where he was an Associate Professor until 2018. From 2015 to 2018, he was a senior computer scientist at the Information Sciences Institute (ISI) and a Visiting Associate Professor at the Institute for Robotics and Intelligent Systems, Viterbi School of Engineering, both at USC, CA, USA, working on the IARPA Janus face recognition project. From 2018 to 2019, Tal was a Principal Applied Scientist at AWS, where he led the design and development of the latest AWS face recognition pipelines. Since June 2019, he is an Applied Research Lead at Facebook AI, where he supported the text (OCR) and people (faces) photo understanding teams. Tal is an associate editor for both IEEE TPAMI and IEEE TBIOM. Some of his recent organizational roles include program chair for WACV'17, ICCV'21, and ECCV'22.
\end{IEEEbiography}

\vfill

\end{document}